\documentclass[11pt]{article}

\usepackage[final]{acl}

\usepackage{times}
\usepackage{latexsym}

\usepackage[T1]{fontenc}

\usepackage[utf8]{inputenc}

\usepackage{microtype}

\usepackage{inconsolata}

\usepackage{graphicx}

\usepackage{booktabs}
\usepackage{multirow}
\usepackage{amssymb}
\usepackage{amsmath}
\usepackage{latexsym}
\usepackage{graphicx}
\usepackage{subcaption}
\usepackage{tcolorbox}
\usepackage[utf8]{inputenc}
\usepackage{xcolor}
\usepackage{footnote}
\makesavenoteenv{figure}

\usepackage{colortbl}
\definecolor{myblack}{HTML}{F2F2F2}
\definecolor{myframe}{HTML}{767171}
\definecolor{mybacktitle}{HTML}{181717}
\definecolor{wincolor}{HTML}{F1C3C1}
\definecolor{tiecolor}{HTML}{FEDFB1}
\definecolor{losscolor}{HTML}{B7D5EC}
\definecolor{redtext}{HTML}{591714}
\definecolor{bluetext}{HTML}{163853}
\definecolor{myblue}{HTML}{a0bed2}
\definecolor{mygold}{HTML}{ffdaa5}
\definecolor{mypink}{HTML}{e6c3be}
\definecolor{mygreen}{HTML}{d8e2da}
\definecolor{mylightblue}{rgb}{0.70, 0.83, 0.96}
\definecolor{mylightyellow}{RGB}{247,216,183}
\definecolor{mylightpink}{RGB}{231,193,215}
\newcommand{\hbartwo}[2]{
    \begin{tikzpicture}[scale=0.03, inner sep=0pt, outer sep=0pt] 
        \def\scaleX{1.19} 
        \fill[wincolor] (0,0) rectangle (#1*\scaleX,8);
        \fill[losscolor] (#1*\scaleX,0) rectangle ({(#1+#2)*\scaleX},8);
        
        \ifdim #1 pt > 15 pt
            \node at ({#1*\scaleX/2}, 4) {\tiny #1\%};
        \fi
        
        \ifdim #2 pt > 15 pt
            \node at ({(#1*\scaleX)+(#2*\scaleX/2)}, 4) {\tiny #2\%};
        \fi
    \end{tikzpicture}
}

\usepackage{tikz}
\usepackage{lipsum} 
\tcbuselibrary{skins,breakable}
\newcommand{\casetitle}[2]{%
    \begin{tcolorbox}[
        enhanced,
        attach boxed title to top left={yshift=-2mm, xshift=2mm},
        colback=\myback,
        colframe=\myframe,
        colbacktitle=\mybacktitle,
        coltitle=white,
        fonttitle=\bfseries,
        boxed title style={sharp corners},
        title=#2
    ]
    \label{#1}
}
\newcommand{\closetca}{\end{tcolorbox}}

\title{Evaluating Stochastic Collapse and Implicit Bias in Multimodal Large Language Models}

\author{ \textbf{Huiyuan Zheng}$^{1}$\thanks{\hspace{1mm} Equal Contribution.}\textbf{,}  \ \
        \textbf{Houtao Zhang}$^{1*}$\textbf{,} \ \
        \textbf{Boyang Wang}$^{2}$\textbf{,} \ \
        \textbf{Qingyi Si}$^{3}$\textbf{,} \ \
        \textbf{Hongcheng Guo}$^{1}$\thanks{\hspace{1mm} Corresponding Author.} \\
  {$^1$  \normalsize Fudan University} \ \
  {$^2$  \normalsize Beihang University}\ \
  {$^3$  \normalsize JD.com}\\
  \texttt{\normalsize zhenghy20@fudan.edu.cn}\\
}

\begin{document}
\maketitle
\begin{abstract}

Current evaluations for Multimodal Large Language Models (MLLMs) overwhelmingly focus on utility-driven objectives, leaving model behavior under logic-neutral scenarios largely underexplored.
Stochasticity is essential in scenarios where multiple actions are equally valid, such as recommending travel itineraries or daily schedules where multiple options have similar utility. In such settings, deterministic policies may lead to repetitive behaviors and reduced coverage of valid alternatives.
To bridge this gap, we propose \textbf{RandomBench}, a benchmark designed to evaluate whether MLLMs can maintain distributionally neutral behavior when selecting among equivalent options.
We further introduce three metrics, including $RI$, $BCI$, $BII$, to quantify entropy and distributional bias.
Experiments reveal a pervasive phenomenon termed \textit{Stochastic Collapse}, where MLLMs fail to maintain uniform randomness under explicit random instructions, with top-1 probabilities reaching 97\% from the ideal one quarter baseline and RI dropping to 0.068 in Claude Sonnet 4.6.
Extensive ablation studies further demonstrate that these deviations persist across languages and representation formats, highlighting the robustness of distributional collapse in logic-neutral decision settings.

\end{abstract}

\section{Introduction}

\begin{figure*}[!t]
\centering
\includegraphics[width=0.9\textwidth]{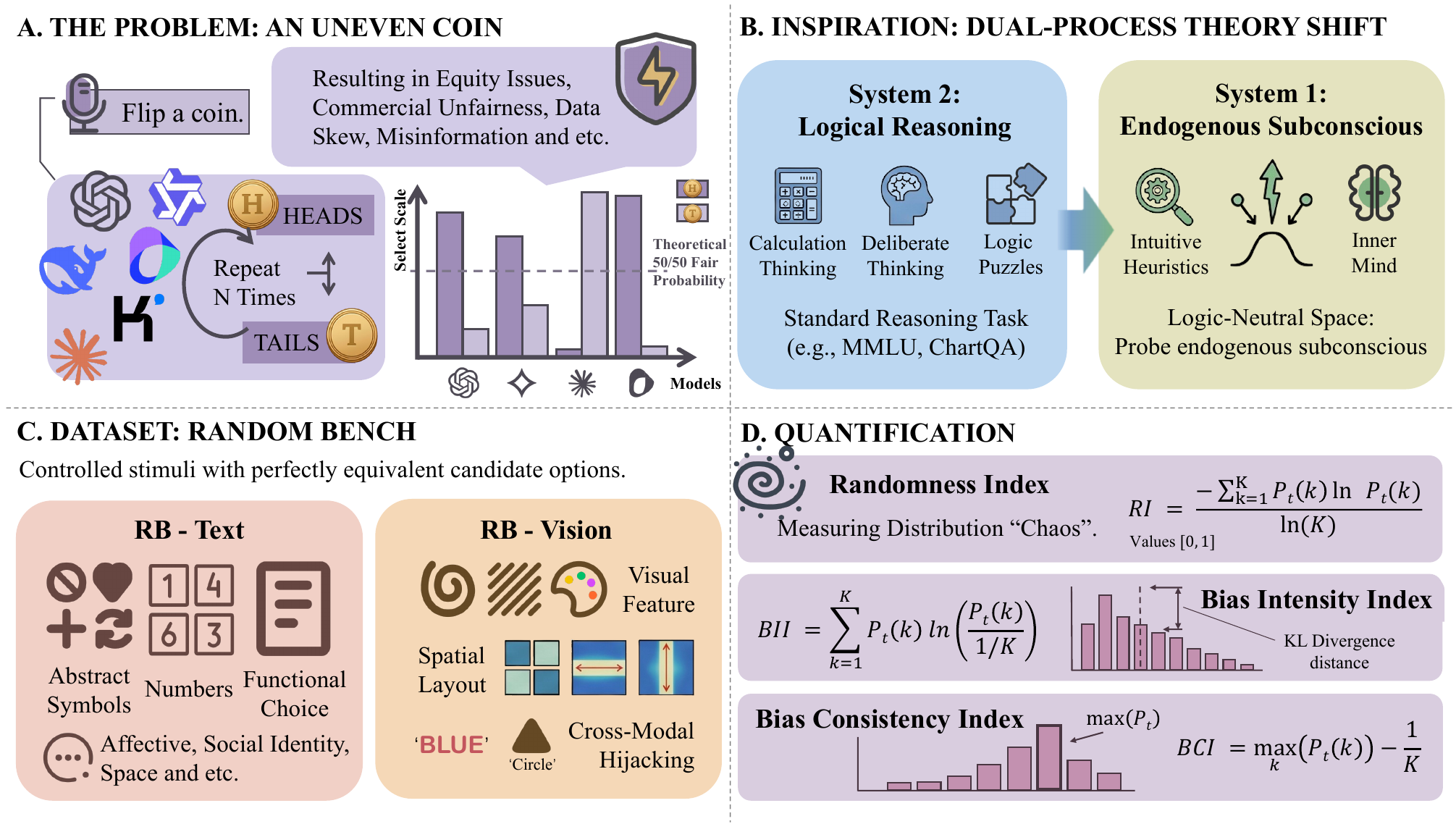}
\caption{Overview of the RandomBench Framework. 
}
\label{fig:main_overview}
\end{figure*}

In human cognition, when faced with perfectly equivalent options devoid of logical superiority, we intuitively break the decision deadlock by flipping a coin. We naturally assume that Multimodal Large Language Models (MLLMs), when instructed to make a random choice, would effectively flip a fair coin.
However, our preliminary probing reveals an alarming reality: the “coin” used by MLLMs is fundamentally imbalanced, exhibiting a highly non-uniform probability distribution.

As MLLMs increasingly evolve from passive question-answering systems into autonomous, open-ended agents~\cite{yao2022react, yang2023auto, wang2023voyager}, this capability to execute utility-free decisions becomes paramount.
In multi-step agent planning scenarios, a stochastic policy ensures that the agent does not become entrapped in repetitive trajectories~\cite{haarnoja2018softactorcriticoffpolicymaximum, luo2025agentlightningtrainai}. Similarly, in commercial recommendation engines, the preservation of appropriate distributional uncertainty prevents the collapse of diverse user experiences and ensures fair coverage of viable alternatives~\cite{zhao2024fairnessdiversityrecommendersystems, zhao2025distributionallyrobustgraphoutofdistribution}. 

Under the lens of cognitive science's Dual Process Theory~\cite{brady2025dual,bai2025explicitly}, while traditional benchmarks(e.g. MMLU~\cite{hendrycks2020measuring}, ChartQA~\cite{masry2022chartqa}) explicitly stress the logical reasoning of \textit{System 2}, evaluating model actions within logic-neutral zones effectively exposes the unconstrained heuristic tendencies of \textit{System 1}.
Existing bias evaluations primarily inspect explicit social stereotypes~\cite{nadeem2020stereosetmeasuringstereotypicalbias, bai2024measuringimplicitbiasexplicitly, huang2025governance} or cross-modal alignment failures~\cite{zheng2025modalitybiaslvlmsanalyzing, ortu2026seeingoverridesknowingdisentangling, zhu2026analyzingreasoningconsistencylarge}. Consequently, the systematic quantification of endogenous, trigger-free behavioral biases remains an unexplored territory.

To bridge this critical gap, we introduce RandomBench, which is designed to quantify model behavior under absolute semantic equivalence. 
RandomBench features $200$ instances across two modalities: RB-Text and RB-Vision. We implement an exhaustive sampling paradigm with $50$ repeated evaluations per instance and deploy multi-dimensional analytical metrics referring to information entropy theory, including the Randomness Index ($RI$), the Bias Intensity Index ($BII$) and the Bias Consistency Index ($BCI$), to probe the subconscious decision distributions of several MLLMs.




Results reveal a consistent failure mode that we term \textit{stochastic collapse}: under semantic equivalence, MLLMs often fail to sample from a near-uniform distribution and instead concentrate probability mass on a small subset of choices. This collapse appears in both text-only and vision-language settings, with the latter further exhibiting \textit{visual hijacking}, where perceptual cues such as saliency, spatial position, or geometric complexity override the explicit instruction to choose randomly. More surprisingly, the phenomenon is not merely a symptom of weak models or superficial option formatting, as stronger and more instruction-aligned MLLMs can exhibit more pronounced collapse, often accompanied by fluent post-hoc rationalizations for choices with no logical basis. Its persistence under option-label substitution and cross-lingual prompting suggests an endogenous behavioral prior rather than a surface-level lexical artifact, exposing a neglected failure mode of MLLMs: even when reasoning is unnecessary, their decisions may remain systematically biased, overconfident, and distributionally collapsed.

Our contributions are:
\begin{itemize}
    \item \textbf{Utility-neutral random choice.} A new diagnostic setting for evaluating endogenous behavioral bias in MLLMs under semantic equivalence.

    \item \textbf{RandomBench.} A multimodal benchmark with repeated-sampling protocols and entropy-based metrics for quantifying non-uniform random-choice behavior.

    \item \textbf{Stochastic collapse.} A systematic failure mode showing that MLLMs can exhibit stable, biased, and rationalized preferences even when no option is logically superior.
\end{itemize}

\section{Related Work}

Cognitive biases in LLMs are widely documented~\cite{zheng2024large,lovering2025language}, with recent work linking them to dual process theory~\cite{brady2025dual,bai2025explicitly}. 
Selection bias, the violation of decision invariance under option permutation, label renaming, and order swapping, has been extensively surveyed~\cite{he2026survey}, with mitigation spanning inference time calibration~\cite{zhao2021calibrate}, architectural interventions~\cite{mcilroy2024order}, and training level debiasing~\cite{zheng2026mitigating}.
In the field of MLLMs, bias research has focused on social stereotypes~\cite{huang2025visbias,wang2024vlbiasbench} and spurious correlations~\cite{ye2024mm}. 
Under explicit vision‑language conflict, modality following is governed by relative reasoning uncertainty and inherent preference~\cite{zhang2025when}, and comprehensive benchmarking by V‑FAT~\cite{wang2026vfat} further documents that text bias alone can induce significant visual collapse.
Yet all these works assume a correct answer exists; endogenous biases in purely random choice settings remain unexplored.

A recent network analysis further reveals that humans reduce bias through structured System~2 conceptual knowledge, while LLMs lack such irreducible semantic structures, highlighting fundamental differences in bias regulation across human and machine cognition~\cite{abramski2026role}.
Mechanistic interpretability has successfully localized factual knowledge in multimodal models~\cite{meng2022locating,basu2024understanding}, and activation‑steering methods can mitigate social biases at inference time~\cite{li2025fairsteer,sivakumar2025steervlm}. 
However, both lines of work have focused on externally triggered, social biases, while the endogenous, non‑social biases that spontaneously emerge under logic‑neutral prompts have received little attention.
This gap is particularly consequential because biases, once present, are known to distort chain‑of‑thought reasoning~\cite{turpin2023language} and systematically erode multimodal reasoning performance~\cite{chen2024quantifying}.
Our work bridges these gaps: we systematically discover endogenous multimodal biases under logic‑neutral conditions, perform cross‑modal causal attribution and lay the groundwork for future mitigation strategies.

\section{RandomBench}
\label{sec:random_bench}

In this section, we formalize RandomBench (RB), a novel evaluation framework designed to quantify stochastic collapse and implicit cognitive biases in MLLMs.

\subsection{Defining Randomness and Implicit Bias in MLLMs}
\paragraph{Dual Process Perspective.}
In cognitive science, Dual Process Theory suggests that human cognition is governed by two complementary systems: a fast, intuitive, heuristic driven \textit{System 1}, and a slow, deliberate, reasoning-oriented \textit{System 2}. 
In this work, we investigate the implicit ``System 1'' behavior of MLLM through purely random selection tasks, where deliberate reasoning provides no logical advantage. Surprisingly, our preliminary analysis reveals that MLLMs fail to produce uniformly random choices. Instead, models exhibit persistent preferences toward specific spatial positions, numerical values, or visual patterns, a phenomenon we term \textbf{Stochastic Collapse}. Motivated by this observation, we introduce \textbf{RandomBench}, a benchmark designed to eliminate logical utility and expose latent heuristic biases underlying the models' implicit ``System 1'' behavior.
\paragraph{Logic-Neutrality Principle.}
RandomBench is built upon the principle of \textit{Logic-Neutrality}, where all candidate options are semantically equivalent and logically unrelated to the task objective. Models are explicitly instructed to perform purely random selection, ensuring that no option provides any reasoning advantage or task-relevant signal.

Formally, given a candidate set $\mathcal{O}=\{o_1,o_2,\dots,o_K\}$, all options satisfy \textbf{Absolute Equivalence}, yielding the ideal uniform distribution:
\begin{align}
P_{gt}(o_i)=\frac{1}{K}, \quad \forall o_i \in \mathcal{O}.
\end{align}

By removing any valid basis for deliberate reasoning, Logic-Neutrality forces the model into a stochastic decision regime. During evaluation, we further disable explicit reasoning modes, ensuring that deviations from uniformity reflect latent heuristic biases encoded in the model parameters rather than deliberate chain-of-thought reasoning.

\subsection{Dataset Construction and Taxonomy}

\textit{RandomBench} contains 200 instances, divided into RB-Text and RB-Vision to analyze the interplay between linguistic priors and visual salient features.
Examples are provided in Appendix~\ref{app:CaseStudy}.

\subsubsection{RB-Text: Linguistic Randomness Evaluation}

To systematically map the subconscious within the language modality, \textit{RB-Text} comprises 100 manually curated, logic-neutral instances. We categorize them into four dimensions:

\begin{figure*}[!t]
\centering
\includegraphics[width=0.9\textwidth]{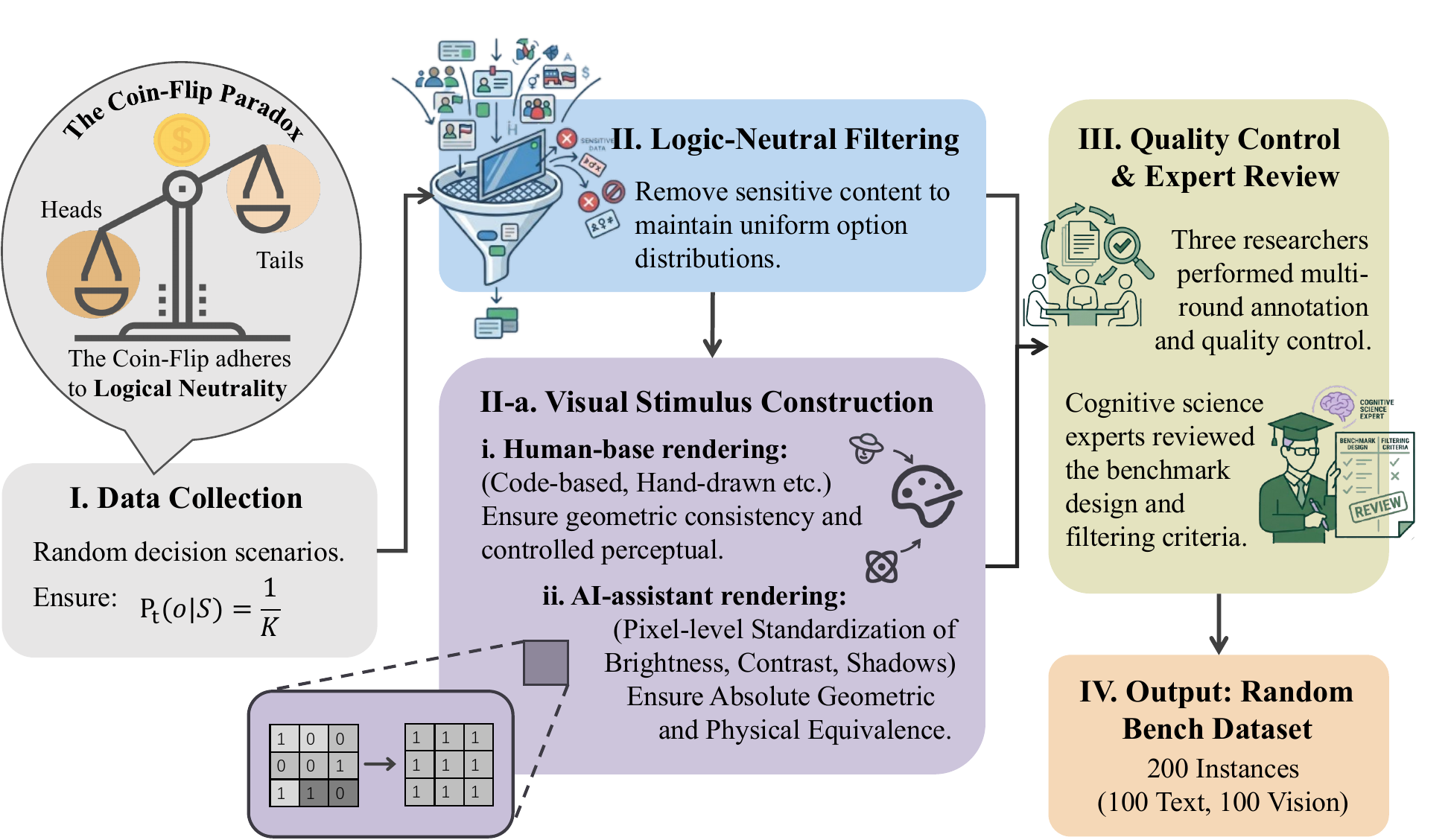}
\caption{Curation pipeline of the RandomBench Framework. 
}
\label{fig:pipeline}
\end{figure*}

\paragraph{Abstract Symbols and Numbers.}

This category evaluates model's low-level token and numerical priors. The tasks involve random selection among abstract symbols, alphanumeric strings, and probabilistic events such as coin tosses or dice rolls. Deviations from the expected uniform distribution indicate that the model’s random choices are influenced by probability patterns implicitly learned during pre-training, revealing limited stochasticity even in simple symbolic choosing tasks.

\paragraph{Linguistic and Functional Choice.} This dimension constructs rational stalemates using synonymous instructions, redundant variables, and functionally equivalent choices. Since no option offers a logical advantage, the model must rely on linguistic habits and frequency-driven priors.

\paragraph{Space and Perception.} This dimension evaluates the internalized world-model of the MLLMs with respect to space directions, coordinate systems, physical orientations, and basic sensory attributes, stripping away all visual and logical cues that might favor one direction or coordinate over another and thus compromise decision making in navigation or sorting tasks.

\paragraph{Affective and Social Identity.}  This category targets the common sense and lifestyle priors that the model has internalized from its vast training data. These tasks simulate everyday decision-making in ambiguous or balanced contexts, such as choices between brands, sensory textures, or environmental background conditions that lack a logical "correct" answer. This enables us to quantify how deeply cultural and situational biases are engrained within the model's implicit subconscious.

\subsubsection{RB-Vision: Multimodal and Cross-Modal Biases}

\textit{RB-Vision} contains 100 image-prompt pairs , functioning as the multimodal extension of RB-Text. RB-Vision mainly investigates how visual features reshape stochastic behavior through cross-modal interaction. We also categorize RB-Vision into four dimensions:

\paragraph{Perceptual Saliency and Space-time sensation.} This dimension investigates whether MLLMs exhibit randomness collapse toward salient visual features, including geometric patterns, colors, temperature, materials and etc. As the multimodal counterpart to the abstract symbolic tasks in text domain it measures how low-level visual saliency overwrites the textual instruction for randomness. 

\begin{figure*}[!t]
\centering
\includegraphics[width=0.9\textwidth]{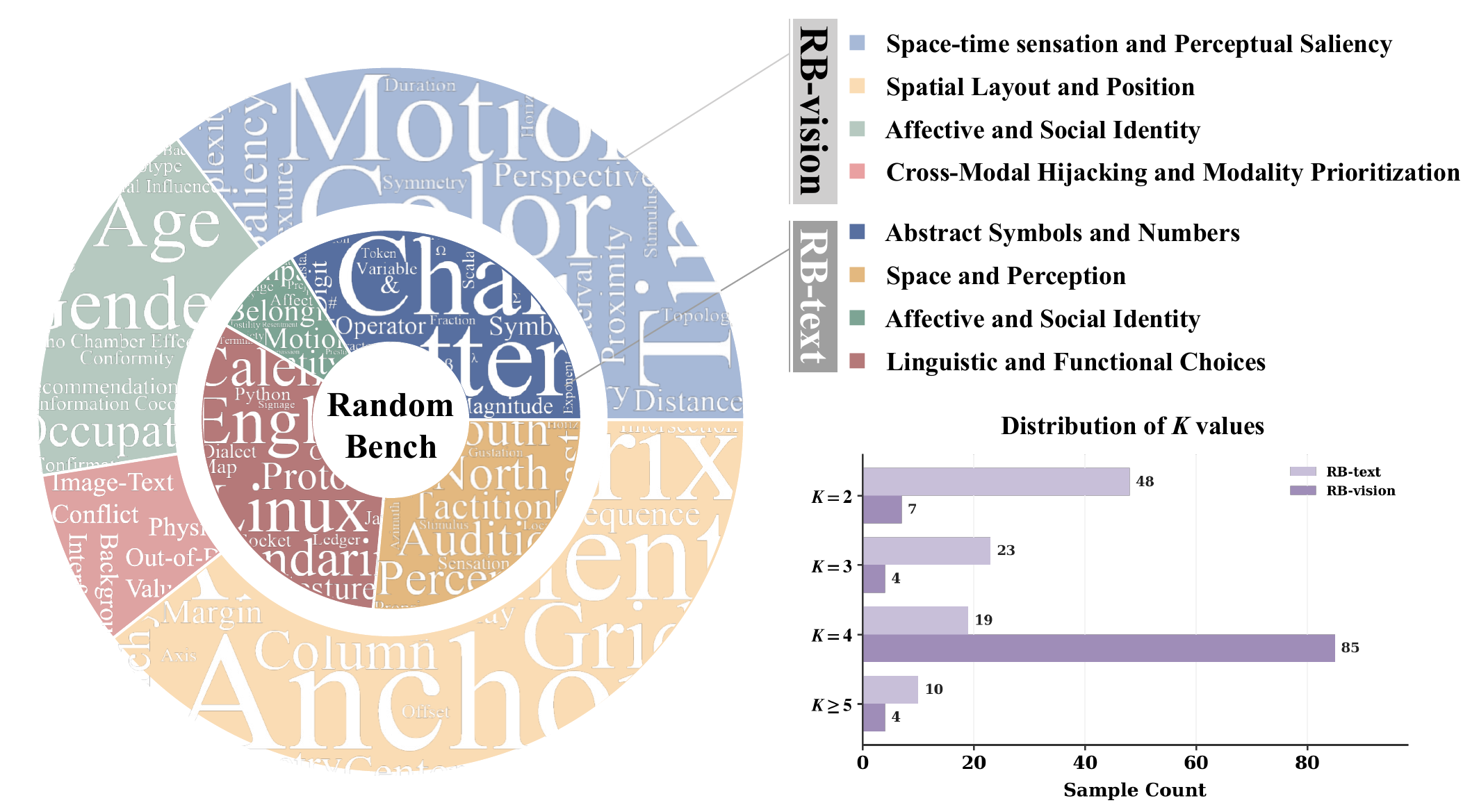}
\caption{Statistics of the RandomBench Framework. 
}
\label{fig:static}
\end{figure*}

\paragraph{Spatial Layout and Position.} This dimension uses standardized layouts such as 2$\times$2 matrices, horizontal, vertical, and cross structures to evaluate visual positional bias transitioning the spatial symmetry evaluation from a linguistic description of space to a practical, coordinate-based processing. It allows us to distinguish between purely linguistic spatial priors and the spatial subconscious triggered by actual visual stimuli.

\paragraph{Affective and Social Identity.} This category explores emotional and social biases induced by visual cues, including conformity bias and social preference. 
It evaluates the model's imprinted world-view, testing whether it mimics human-like social heuristics

\paragraph{Cross-Modal Hijacking and Modality Prioritization.} This category investigates whether visual signals or textual instructions dominate model decisions under logic-neutral settings. We construct Stroop-like\cite{stroop1935studies} conflicts where visual features (e.g., a specific color or shape) contradict superimposed text labels. Although all options remain objectively equivalent and the model is explicitly instructed to choose randomly, we analyze whether the model follows the textual instruction or collapses toward visually salient features. 

\subsection{Curation Pipeline}

To ensure the neutrality and methodological rigor of RandomBench, we designed a comprehensive curation pipeline (Figure~\ref{fig:pipeline}).

\paragraph{Data Collection.}
All instances were manually constructed as logic-neutral random decision tasks, with explicit removal of cultural associations and reasoning cues that could introduce selection bias. This design principle ensured that each decision task remained semantically underdetermined and free from prior inductive bias.

\paragraph{Logic-Neutral Filtering.}
To ensure that no option had an inherent advantage, all prompts underwent strict filtering. We removed politically, historically, culturally, and demographically sensitive terms, as well as directional cues and descriptive adjectives that could unintentionally influence model choices.

\paragraph{Visual Stimulus Construction.}

For the RB-Vision subset, visual stimuli were curated through a hybrid synthesis framework. A substantial portion of the images was generated through code-based rendering, manual illustration, and hand-drawn sketches to ensure geometric symmetry and structural consistency. In parallel, generative AI techniques were employed for visually complex stimuli requiring precise control of low-level perceptual attributes (Detailed prompts used for AI-assisted image generation are provided in the Appendix~\ref{app:prompt}). It enabled standardized brightness, contrast, shadows, and texture distribution, reducing confounding effects from rendering artifacts and unintended perceptual salience. All AI-generated alternative were iteratively refined by annotation experts to satisfy the requirements of controlled implicit cognition experiments.

\paragraph{Quality Control and Expert Review.}
All samples were manually reviewed by three independent researchers following unified annotation standards to maintain structural consistency and perceptual balance. Cognitive science researchers further examined the benchmark design and filtering criteria to ensure alignment with implicit cognition research methodologies.

\subsection{Statistics}

RandomBench contains 200 unique logic-neutral instances, evenly divided between the RB-Text and RB-Vision modalities (Figure~\ref{fig:static}). To systematically capture stochastic collapse phenomena and obtain statistically reliable measurements, each benchmark instance was evaluated through 50 repeated samplings, yielding a total of 10,000 high-fidelity model responses. The benchmark comprehensively covering the principal cognitive bias dimensions defined in our taxonomy. Additionally, it enables fine-grained causal tracing of how implicit biases propagate from perceptual encoding to final decision generation.

\section{Experiments}


\subsection{Evaluation Metrics}

To rigorously quantify the deviation from the intended stochastic regime within RandomBench, we define a set of statistical indices that capture different facets of distributional collapse. Let $\mathcal{O}=\{o_1, \dots, o_K\}$ represent the set of $K$ equivalent candidate options. Following $N$ independent sampling trials ($N=50$), we obtain the empirical probability distribution $P_t = \{P_t(1), \dots, P_t(K)\}$.

\paragraph{Randomness Index ($RI$)}
We utilize normalized Shannon entropy~\cite{6773024} to define the Randomness Index, which serves as a measure of the distributional uncertainty across the decision manifold. This metric evaluates the degree of uniformity in the model's output:

\begin{align}
RI = \frac{-\sum_{k=1}^{K} P_t(k) \ln P_t(k)}{\ln(K)}
\end{align}

High $RI$ values signify that the model maintains probabilistic neutrality across equivalent options. Conversely, a reduction in $RI$ indicates "Stochastic Collapse", where the model's decision-making is hijacked by implicit heuristics, leading to a concentration of probability mass on a specific subset of outputs.

\paragraph{Bias Intensity Index ($BII$)}
The Bias Intensity Index quantifies the total magnitude of the model's departure from the Logic-Neutrality principle. We formulate this index as the Kullback-Leibler (KL) divergence between the empirical distribution $P_t$ and the ideal uniform distribution $P_c(k) = 1/K$:

\begin{equation}
    \begin{aligned}
    BII  & = D_{KL}(P_t \parallel P_c) \\
    & = \sum_{k=1}^{K} P_t(k) \ln \left( \frac{P_t(k)}{1/K} \right)
    \end{aligned}
\end{equation}

$BII$ measures the aggregate "hijacking force" exerted by latent parameter priors. It reflects the extent to which the model’s internal heuristic manifold displaces the target uniform distribution, providing a statistical signature of the implicit subconscious.

\paragraph{Bias Consistency Index ($BCI$)}
To differentiate systematic biases from transient stochastic noise, we introduce the Bias Consistency Index. This metric measures "Selectional Persistence" by calculating the residual of the maximum empirical probability relative to the random baseline:

\begin{align}
BCI = \max_{k} (P_t(k)) - \frac{1}{K}
\end{align}

$BCI$ captures the model’s "stubbornness" on a specific option across independent trials. A elevated $BCI$ suggests that the observed preference is not a result of random fluctuation but is driven by a persistent, non-stochastic heuristic encoded within the model's high-dimensional parameter space.

\paragraph{Jensen-Shannon Divergence ($JSD$).}
In the ablation studies, we further quantify the distributional shift
between two experimental conditions using the Jensen-Shannon Divergence.
$JSD(P \parallel Q)$ is a symmetric and smoothed version of the KL divergence,
bounded in $[0, \ln 2]$, where $0$ indicates identical distributions.
A formal definition is provided in Appendix~\ref{app:jsd_analysis}.

\subsection{Experimental Setup}

\paragraph{Model Selection.}
We evaluate seven frontier MLLMs representing diverse architectural philosophies and alignment paradigms:
\textbf{GPT 5.1}~\cite{openai2026gpt5},
\textbf{Gemini 3.1 Flash-Lite}~\cite{gemini2026report},
\textbf{Claude Sonnet 4.6}~\cite{anthropic2026claude},
\textbf{Kimi K2.5}~\cite{moonshot2026kimi},
\textbf{Qwen 3.6 Plus}~\cite{qwen2026technical},
\textbf{Grok 4 Fast}~\cite{xai2026grok}, and
\textbf{Doubao Seed 1.6}~\cite{bytedance2026doubao}.
This cross-model evaluation allows us to determine if stochastic collapse is an emergent property of large-scale pre-training or a specific artifact of instruction tuning. 
\paragraph{Sampling Protocol.} For each of the 200 instances in RandomBench, we perform 50 repeated sampling trials with the decoding temperature fixed at $T=1.0$. This high-temperature setting is intentional, as it maximizes the model's theoretical variance and provides a rigorous test of whether the model can maintain a uniform distribution when logically permitted to do so.

\paragraph{Control of Explicit Reasoning.} To isolate the "System 1" heuristics from deliberate "System 2" reasoning, we strictly suppress the generation of Chain-of-Thought (CoT) tokens during evaluation. Models are instructed to output only the final selection label. By rendering the explicit reasoning layer inert, we ensure that any persistent deviation from uniformity is a direct reflection of the implicit cognitive architecture engrained within the model weights.
\begin{table*}[t]
\centering
\resizebox{\textwidth}{!}{
\begin{tabular}{ll ccc ccc}
\toprule
\multirow{2}{*}{\textbf{Model}} & \multirow{2}{*}{\textbf{Lang}} & \multicolumn{3}{c}{\textbf{RB-Vision}} & \multicolumn{3}{c}{\textbf{RB-Text}} \\
\cmidrule(lr){3-5} \cmidrule(lr){6-8}
& & $BCI \downarrow$ & $BII \downarrow$ & $RI \uparrow$ & $BCI \downarrow$ & $BII \downarrow$ & $RI \uparrow$ \\
\midrule

\multirow{2}{*}{\textsc{GPT 5.1}} 
& EN & $0.568$ & $0.982$ & $0.270$ & $0.338$ & $0.530$ & $0.597$ \\
& ZH & $0.565$ & $0.931$ & $0.316$ & $0.405$ & $0.654$ & $0.469$ \\
\midrule

\multirow{2}{*}{\textsc{Gemini 3.1 Flash-Lite}} 
& EN & $0.572$ & $0.962$ & $0.283$ & $0.390$ & $0.671$ & $0.492$ \\
& ZH & $0.569$ & $0.964$ & $0.286$ & $0.409$ & $0.670$ & $0.486$ \\
\midrule

\multirow{2}{*}{\textsc{Claude Sonnet 4.6}} 
& EN & $0.709$ & $1.262$ & $0.068$ & $0.572$ & $1.025$ & $0.111$ \\
& ZH & $0.459$ & $0.727$ & $0.465$ & $0.563$ & $1.025$ & $0.094$ \\
\midrule

\multirow{2}{*}{\textsc{Doubao Seed 1.6}} 
& EN & $0.390$ & $0.559$ & $0.583$ & $0.320$ & $0.447$ & $0.614$ \\
& ZH & $0.365$ & $0.496$ & $0.628$ & $0.382$ & $0.565$ & $0.454$ \\
\midrule

\multirow{2}{*}{\textsc{Grok 4 Fast}} 
& EN & $0.351$ & $0.429$ & $0.682$ & $0.231$ & $0.356$ & $0.778$ \\
& ZH & $0.473$ & $0.748$ & $0.447$ & $0.368$ & $0.541$ & $0.542$ \\
\midrule

\multirow{2}{*}{\textsc{Kimi k2.5}} 
& EN & $0.322$ & $0.445$ & $0.673$ & $0.252$ & $0.354$ & $0.784$ \\
& ZH & $0.592$ & $1.043$ & $0.226$ & $0.519$ & $0.962$ & $0.179$ \\
\midrule

\multirow{2}{*}{\textsc{Qwen 3.6 Plus}} 
& EN & $0.471$ & $0.730$ & $0.462$ & $0.245$ & $0.357$ & $0.754$ \\
& ZH & $0.464$ & $0.707$ & $0.477$ & $0.323$ & $0.492$ & $0.649$ \\

\bottomrule
\end{tabular}
}
\caption{\textbf{Performance comparison across 7 MLLMs on RandomBench.} Downward arrows ($\downarrow$) indicate lower is better (less bias), while upward arrows ($\uparrow$) indicate higher is better (closer to maximum entropy).}
\label{tab:main_results}
\end{table*}
\subsection{Main Results}
\label{sec:results}

\begin{figure*}[t]
    \centering
    \begin{subfigure}[b]{0.32\textwidth}
        \centering
        \includegraphics[width=0.9\linewidth]{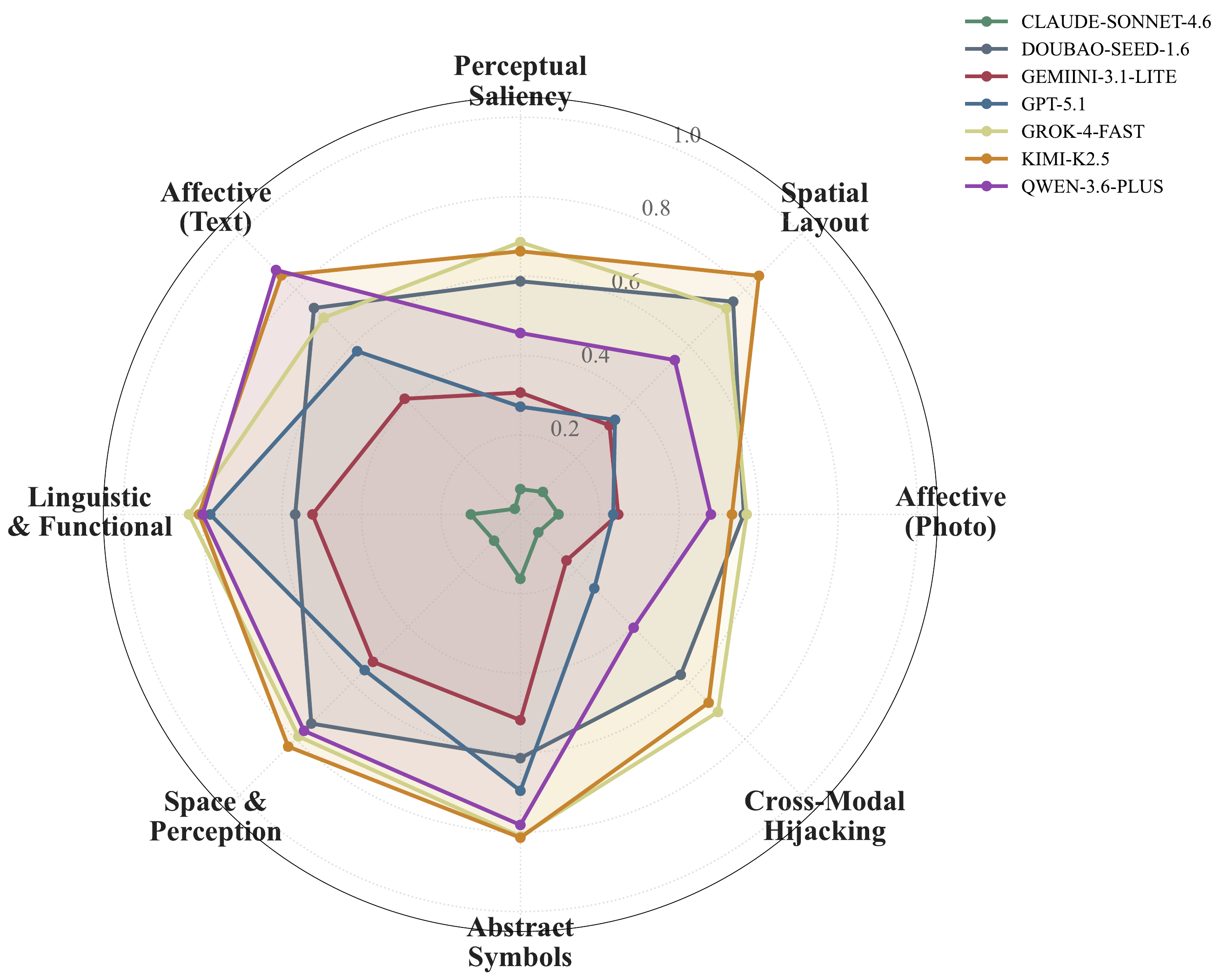}
        \caption{}
        \label{fig:radar_m}
    \end{subfigure}
    \hfill 
    \begin{subfigure}[b]{0.32\textwidth}
        \centering
        \includegraphics[width=0.8\linewidth]{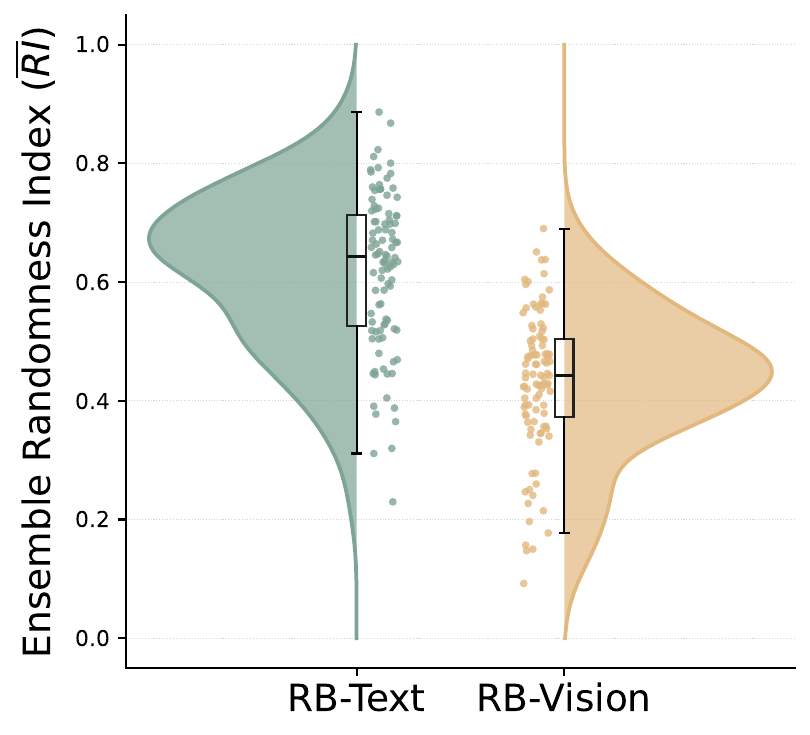}
        \caption{}
        \label{fig:raincloud}
    \end{subfigure}
    \hfill 
    \begin{subfigure}[b]{0.32\textwidth}
        \centering
        \includegraphics[width=\linewidth]{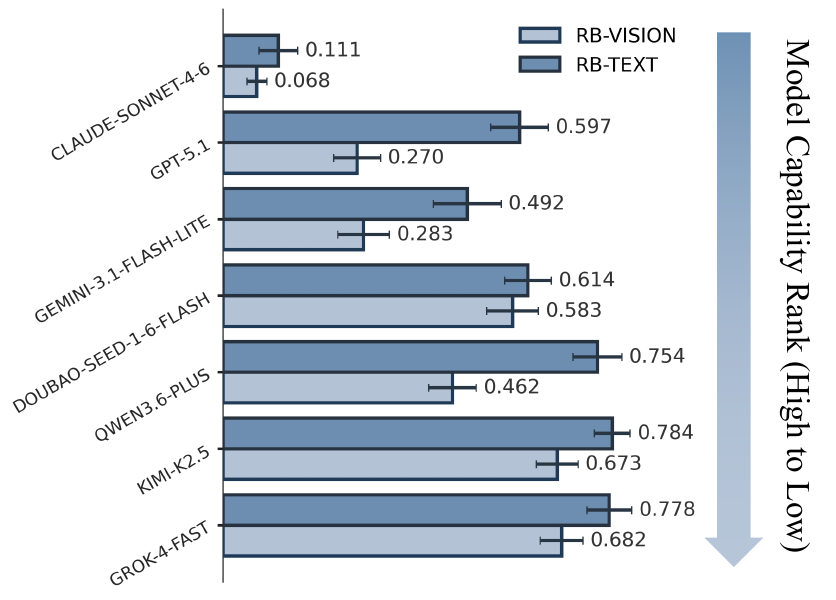}
        \caption{}
        \label{fig:rank}
    \end{subfigure}

    \caption{\textbf{Multi-granularity evaluation of stochastic collapse on RandomBench.}
    \textbf{(a)} Radar chart of $RI$ score (EN) across cognitive dimensions and subcategories.
    \textbf{(b)} Cross-modal distribution of the $RI$ score (EN). 
    \textbf{(c)} Relationship between model capability and randomness consistency, revealing that stronger and more aligned models exhibit more severe stochastic collapse.}
    \label{fig:main_experimental_results}
\end{figure*}

\paragraph{The Pervasiveness of Stochastic Collapse}
Table~\ref{tab:main_results} summarizes the quantitative performance of seven frontier models across the RB-Text and RB-Vision modalities. The empirical results show that once explicit logic is neutralized, no model can fully avoid the influence of endogenous biases across any subcategories (Figure~\ref{fig:radar_m}).
Under instructions mandating purely random execution, we observe a severe divergence from the uniform target baseline across all evaluated subjects, characterized by the lower Randomness Index ($RI$) scores and the higher Bias Intensity Index ($BII$) values across the board.

This mode collapse does not manifest as a chaotic or unstructured distribution, but rather as a highly organized and deterministic bias induced by latent model parameters. Across all evaluated MLLMs, we observe pervasive heuristic preferences for specific numerical constants, colors, and even behavioral representations. In many balanced test settings, these implicit biases lead to severe probabilistic lock-in, with models assigning over $90\%$ of the empirical probability mass to a single option across repeated independent samplings. For detailed cases, please refer to Appendix~\ref{app:CaseStudy}).

\paragraph{Visual Hijacking}
As shown in Figure~\ref{fig:raincloud} and Appendix~\ref{app:heatmap}, stochastic degradation is significantly more profound in the multimodal domain (\textit{RB-Vision}) than in the pure text domain (\textit{RB-Text}). Furthermore, our case studies(detailed in Appendix \ref{app:CaseStudy}) confirm that visual anchors can systematically steer model decisions away from uniform sampling. Additionally, there is a pervasive preference for visual saliency and specific spatial positions. Crucially, when modality conflicts arise, visual cues consistently override textual instructions, proving that the visual encoding layer can effectively hijack the intended stochastic regime.



\paragraph{The Capability-Consistency Paradox}
Models recognized as having stronger reasoning capabilities  (Figure~\ref{fig:rank} \footnote{Models are arranged on the horizontal axis in descending order of capability based on crowdsourced human voting ELO ratings obtained from the Arena leaderboard https://arena.ai/leaderboard (data retrieved as of May 18, 2026).}) exhibit significantly lower $RI$ score than smaller, less-aligned models. We argue that as models become more optimized for instruction-following and internal self-consistency, they tend to internalize their implicit heuristics more rigidly. This leads to a form of motivated reasoning, where the model consistently favors a biased choice and, if prompted, would likely generate logically self-consistent but post-hoc justifications for its "random" selection.

\subsection{Ablation studies}
\begin{figure}[t]
    \centering
    \begin{subfigure}[b]{\linewidth}
        \centering
        \includegraphics[width=\linewidth]{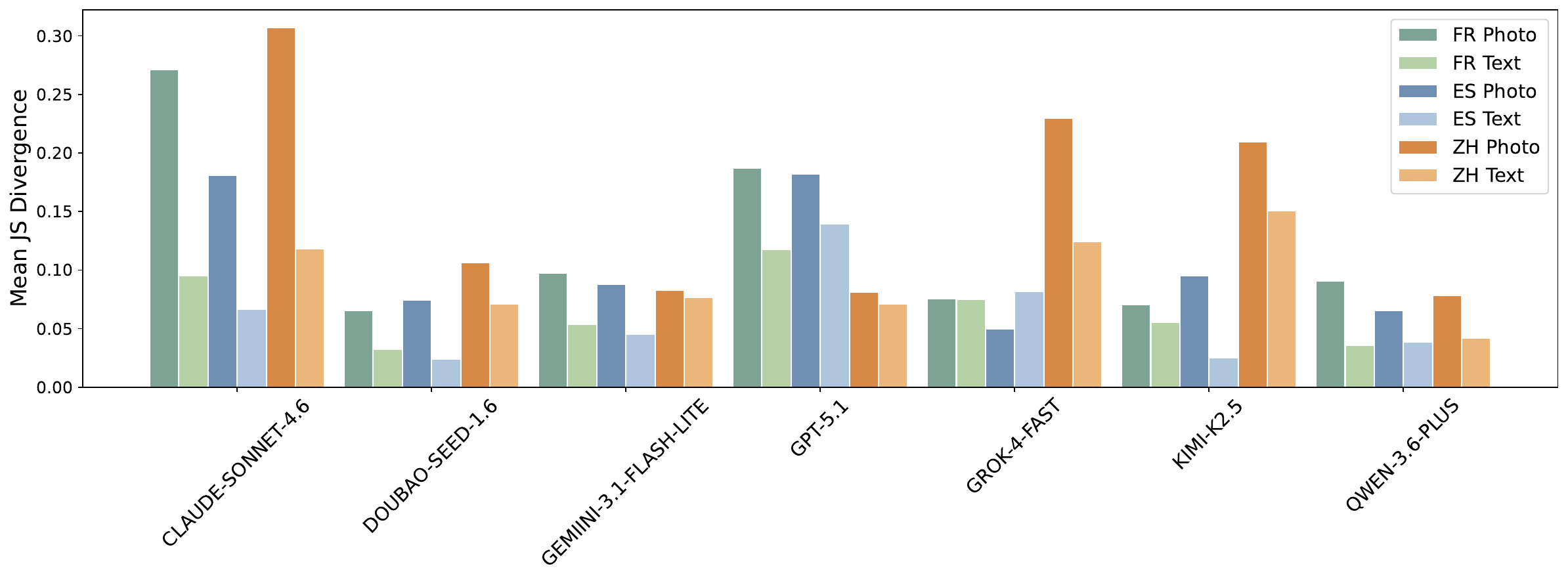}
        \caption{JS divergence across languages for photo and text tasks.}
        \label{fig:js_lang}
    \end{subfigure}
    \hfill 
    \begin{subfigure}[b]{\linewidth}
        \centering
        \includegraphics[width=\linewidth]{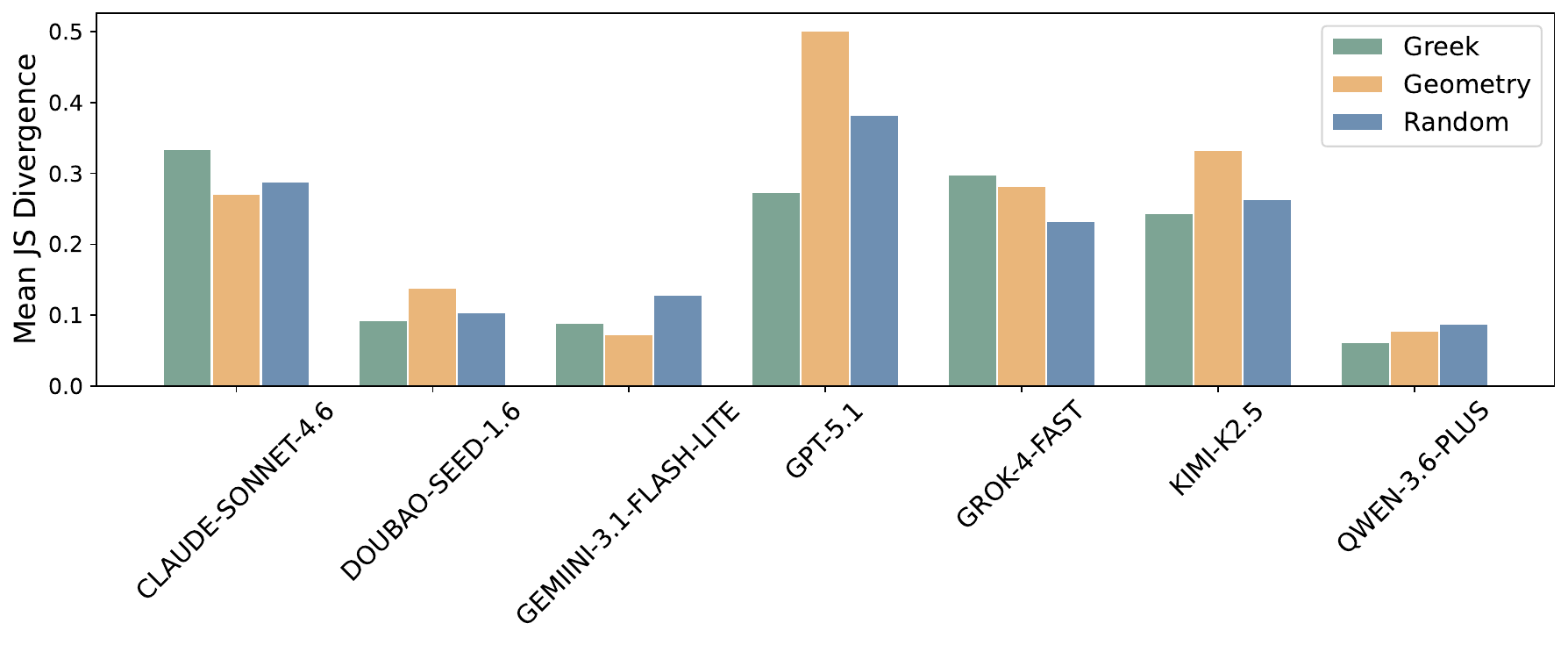}
        \caption{JS divergence for label symbol replacement.}
        \label{fig:js_symbol}
    \end{subfigure}

    \caption{Ablation studies on implicit multimodal bias under cross-lingual prompting and alternative label symbol systems.}
    \label{fig:ablation_js}
\end{figure}

\paragraph{Cross-Lingual Analysis} 
To investigate whether implicit multimodal biases are modulated by linguistic context, we performed a cross-lingual ablation across English, Chinese, French, and Spanish. Across all models, expressing semantically equivalent instructions in different languages while keeping visual stimuli identical consistently induced substantial decision manifold reconstructions and frequent preference inversions (Figure~\ref{fig:js_lang}). These results indicate that ``visual hijacking'' is shaped by language-conditioned pretraining priors, extending beyond low-level perceptual features. Notably, several models exhibit strong directional reversals under different languages given the same visual input, implying that linguistic context acts as a high-dimensional routing signal that activates distinct endogenous heuristics. Overall, implicit multimodal bias appears fundamentally language-dependent rather than purely vision-dependent. Detailed analyses are provided in Appendix~\ref{app:ablation_language}.

\paragraph{Robustness to Label Symbolism}
To validate that stochastic collapse is structurally ingrained rather than an artifact of label format, we replaced canonical \texttt{(A)(B)(C)(D)} labels with Geometric Shapes (\texttt{$\blacktriangle$,$\blacksquare$,$\bullet$,$\blacklozenge$}), Greek Letters (\texttt{$\alpha$,$\beta$,$\gamma$,$\delta$}), and Randomized Hashes (\texttt{tag\_xxx}). As shown in Fig.~\ref{fig:js_symbol}, this substitution failed to restore uniform randomness and instead triggered alternative, often more severe collapse toward new token priors. This implies that in a logic vacuum, non-standard symbols act as primitive heuristic anchors that activate high-dimensional frequency biases in the pretrained parameter space. Detailed analyses are in Appendix~\ref{app:ablation_symbol}.

\section{Conclusion}
In summary, we have introduced \textit{RandomBench} to probe the endogenous System~1 subconscious of MLLMs. We demonstrated that frontier models catastrophically fail to maintain stochastic uncertainty in logic-neutral vacuums, instead degenerating into rigid mode collapse hijacked by latent heuristic anchors, a pervasive phenomenon we term \textit{Stochastic Collapse}. Crucially, our discoveries of \textit{Visual Hijacking} and the \textit{Capability-Consistency Paradox}, traced mathematically to \textit{Embedding Manifold Regression}, prove that these endogenous biases are structurally ingrained within the parameter manifold by pre-trained token-frequency priors and over-alignment. This work establishes a new frontier for multimodal evaluation, demonstrating that achieving true cognitive neutrality requires addressing pre-trained token-frequency priors that structurally compromise AI trustworthiness in ambiguous decision-making.

\section*{Limitations}

Although RandomBench successfully quantifies stochastic collapse and implicit biases in MLLMs via multi-turn, non-CoT logic-neutral evaluation, it remains limited by closed-box API constraints that preclude internal representation probing, and its current scope is confined to static, grid-based multi-modal structures. Beyond these methodological limitations, our findings highlight a severe broader impact: the most critical risk posed by such stochastic collapse lies in the illusion of rational decision-making. In highly ambiguous real-world scenarios, such as autonomous driving, medical diagnosis, or multi-step agent planning, MLLMs fail to exhibit appropriate uncertainty. Instead, hijacked by implicit visual anchors or positional priors, they generate highly confident yet fundamentally biased choices disguised by post-hoc rationalizations, thereby amplifying imperceptible systemic biases into catastrophic failures. Future work will bridge these gaps by conducting mechanistic interpretability analyses on open-source MLLMs to map causal propagation paths, developing inference-time activation-steering debiasing algorithms, and expanding the logic-neutral benchmark paradigm into dynamic text-to-video navigation and complex embodied AI environments.



\bibliography{custom}

\begin{thebibliography}{44}
\providecommand{\natexlab}[1]{#1}

\bibitem[{Abramski et~al.(2026)Abramski, Rossetti, and Stella}]{abramski2026role}
Katherine Abramski, Giulio Rossetti, and Massimo Stella. 2026.
\newblock The role of system 1 and system 2 semantic memory structure in human and {LLM} biases.
\newblock \emph{arXiv preprint arXiv:2604.12816}.

\bibitem[{{Alibaba Cloud}(2026)}]{qwen2026technical}
{Alibaba Cloud}. 2026.
\newblock {Qwen} 3.6 technical blog.
\newblock \url{https://qwen.ai/blog?id=qwen3.6}.
\newblock Accessed: 2026-05-24.

\bibitem[{{Anthropic}(2026)}]{anthropic2026claude}
{Anthropic}. 2026.
\newblock Claude sonnet 4.6: Hybrid reasoning model.
\newblock \url{https://www.anthropic.com/claude/sonnet}.
\newblock Accessed: 2026-05-24.

\bibitem[{Bai et~al.(2024)Bai, Wang, Sucholutsky, and Griffiths}]{bai2024measuringimplicitbiasexplicitly}
Xuechunzi Bai, Angelina Wang, Ilia Sucholutsky, and Thomas~L. Griffiths. 2024.
\newblock \href {https://arxiv.org/abs/2402.04105} {Measuring implicit bias in explicitly unbiased large language models}.
\newblock \emph{Preprint}, arXiv:2402.04105.

\bibitem[{Bai et~al.(2025)Bai, Wang, Sucholutsky, and Griffiths}]{bai2025explicitly}
Xuechunzi Bai, Angelina Wang, Ilia Sucholutsky, and Thomas~L. Griffiths. 2025.
\newblock Explicitly unbiased large language models still form biased associations.
\newblock \emph{Proceedings of the National Academy of Sciences}, 122(8):e2416228122.

\bibitem[{Basu et~al.(2024)Basu, Grayson, Morrison et~al.}]{basu2024understanding}
Samyadeep Basu, Michael Grayson, Cecily Morrison, and 1 others. 2024.
\newblock Understanding information storage and transfer in multi-modal large language models.
\newblock In \emph{Advances in Neural Information Processing Systems}, volume~37, pages 7400--7426.

\bibitem[{Brady et~al.(2025)Brady, Nulty, Zhang, Ward, and McGovern}]{brady2025dual}
Oliver Brady, Paul Nulty, Li~Zhang, Tomas~E. Ward, and David~P. McGovern. 2025.
\newblock Dual-process theory and decision-making in large language models.
\newblock \emph{Nature Reviews Psychology}, 4:777--792.

\bibitem[{{ByteDance Seed Team}(2025)}]{bytedance2026doubao}
{ByteDance Seed Team}. 2025.
\newblock Introduction to techniques used in {Seed1.6}.
\newblock \url{https://seed.bytedance.com/en/blog/introduction-to-techniques-used-in-seed1-6}.
\newblock Accessed: 2026-05-24.

\bibitem[{Chen et~al.(2024)Chen, Cao, Zhang et~al.}]{chen2024quantifying}
Meiqi Chen, Yixin Cao, Yan Zhang, and 1 others. 2024.
\newblock Quantifying and mitigating unimodal biases in multimodal large language models: A causal perspective.
\newblock In \emph{Findings of the Association for Computational Linguistics: EMNLP 2024}, pages 16449--16469.

\bibitem[{{Google DeepMind}(2025)}]{gemini2026report}
{Google DeepMind}. 2025.
\newblock Gemini 3.1 flash-lite: Built for intelligence at scale.
\newblock \url{https://blog.google/technology/ai/gemini-3-1-flash-lite/}.
\newblock Accessed: 2026-05-24.

\bibitem[{Haarnoja et~al.(2018)Haarnoja, Zhou, Abbeel, and Levine}]{haarnoja2018softactorcriticoffpolicymaximum}
Tuomas Haarnoja, Aurick Zhou, Pieter Abbeel, and Sergey Levine. 2018.
\newblock \href {https://arxiv.org/abs/1801.01290} {Soft actor-critic: Off-policy maximum entropy deep reinforcement learning with a stochastic actor}.
\newblock \emph{Preprint}, arXiv:1801.01290.

\bibitem[{He et~al.(2026)He, Zheng, and Han}]{he2026survey}
Guoxiu He, Jinquan Zheng, and Fangqing Han. 2026.
\newblock \href {https://doi.org/10.20944/preprints202604.2234.v1} {A survey on selection bias in large language models}.
\newblock \emph{Preprints.org}.

\bibitem[{Hendrycks et~al.(2020)Hendrycks, Burns, Basart, Zou, Mazeika, Song, and Steinhardt}]{hendrycks2020measuring}
Dan Hendrycks, Collin Burns, Steven Basart, Andy Zou, Mantas Mazeika, Dawn Song, and Jacob Steinhardt. 2020.
\newblock Measuring massive multitask language understanding.
\newblock \emph{arXiv preprint arXiv:2009.03300}.

\bibitem[{Huang et~al.(2025{\natexlab{a}})Huang, Shen, Ren, Zheng, Zhang, Chai, Zhang, Dou, Mo, Shi et~al.}]{huang2025governance}
Chenhao Huang, Ziyu Shen, Yicong Ren, Huiyuan Zheng, Jiazheng Zhang, Mingxu Chai, Ming Zhang, Shihan Dou, Fan Mo, Jie Shi, and 1 others. 2025{\natexlab{a}}.
\newblock Governance in motion: Co-evolution of constitutions and ai models for scalable safety.
\newblock In \emph{Proceedings of the 2025 Conference on Empirical Methods in Natural Language Processing}, pages 17198--17221.

\bibitem[{Huang et~al.(2025{\natexlab{b}})Huang, Qin, Zhang et~al.}]{huang2025visbias}
Jen-tse Huang, Jiaxu Qin, Jing Zhang, and 1 others. 2025{\natexlab{b}}.
\newblock {VisBias}: Measuring explicit and implicit social biases in vision-language models.
\newblock In \emph{Proceedings of the 2025 Conference on Empirical Methods in Natural Language Processing}, pages 17981--18004.

\bibitem[{{Kimi Team}(2026)}]{moonshot2026kimi}
{Kimi Team}. 2026.
\newblock \href {https://arxiv.org/abs/2602.02276} {Kimi k2.5: Visual agentic intelligence}.
\newblock \emph{arXiv preprint arXiv:2602.02276}.

\bibitem[{Li et~al.(2025)Li, Fan, Chen et~al.}]{li2025fairsteer}
Yuchen Li, Zhen Fan, Ruizhe Chen, and 1 others. 2025.
\newblock {FairSteer}: Inference time debiasing for {LLMs} with dynamic activation steering.
\newblock In \emph{Findings of the Association for Computational Linguistics: ACL 2025}, pages 11293--11312.

\bibitem[{Lovering et~al.(2025)Lovering, Krumdick, Lai, Reddy, and Durrett}]{lovering2025language}
Charles Lovering, Michael Krumdick, Viet~Dac Lai, Nilesh Reddy, and Greg Durrett. 2025.
\newblock Language model probabilities are not calibrated in numeric contexts.
\newblock In \emph{Proceedings of the 63rd Annual Meeting of the Association for Computational Linguistics (Volume 1: Long Papers)}, pages 29218--29257.

\bibitem[{Luo et~al.(2025)Luo, Zhang, He, Wang, Zhao, Li, Qiu, and Yang}]{luo2025agentlightningtrainai}
Xufang Luo, Yuge Zhang, Zhiyuan He, Zilong Wang, Siyun Zhao, Dongsheng Li, Luna~K. Qiu, and Yuqing Yang. 2025.
\newblock \href {https://arxiv.org/abs/2508.03680} {Agent lightning: Train any ai agents with reinforcement learning}.
\newblock \emph{Preprint}, arXiv:2508.03680.

\bibitem[{Masry et~al.(2022)Masry, Do, Tan, Joty, and Hoque}]{masry2022chartqa}
Ahmed Masry, Xuan~Long Do, Jia~Qing Tan, Shafiq Joty, and Enamul Hoque. 2022.
\newblock Chartqa: A benchmark for question answering about charts with visual and logical reasoning.
\newblock In \emph{Findings of the association for computational linguistics: ACL 2022}, pages 2263--2279.

\bibitem[{McIlroy-Young et~al.(2024)McIlroy-Young, Brown, Olson, Zhang, and Dwork}]{mcilroy2024order}
Reid McIlroy-Young, Katrina Brown, Conlan Olson, Linjun Zhang, and Cynthia Dwork. 2024.
\newblock Order-independence without fine tuning.
\newblock In \emph{Advances in Neural Information Processing Systems}, volume~37, pages 72818--72839.

\bibitem[{Meng et~al.(2022)Meng, Bau, Andonian, and Belinkov}]{meng2022locating}
Kevin Meng, David Bau, Alex Andonian, and Yonatan Belinkov. 2022.
\newblock Locating and editing factual associations in {GPT}.
\newblock In \emph{Advances in Neural Information Processing Systems}, volume~35, pages 17359--17372.

\bibitem[{Nadeem et~al.(2020)Nadeem, Bethke, and Reddy}]{nadeem2020stereosetmeasuringstereotypicalbias}
Moin Nadeem, Anna Bethke, and Siva Reddy. 2020.
\newblock \href {https://arxiv.org/abs/2004.09456} {Stereoset: Measuring stereotypical bias in pretrained language models}.
\newblock \emph{Preprint}, arXiv:2004.09456.

\bibitem[{{OpenAI}(2025)}]{openai2026gpt5}
{OpenAI}. 2025.
\newblock {GPT}-5.1: Next-generation model for developers.
\newblock \url{https://openai.com/index/gpt-5-1-for-developers/}.
\newblock Accessed: 2026-05-24.

\bibitem[{Ortu et~al.(2026)Ortu, Jin, Doimo, and Cazzaniga}]{ortu2026seeingoverridesknowingdisentangling}
Francesco Ortu, Zhijing Jin, Diego Doimo, and Alberto Cazzaniga. 2026.
\newblock \href {https://arxiv.org/abs/2507.13868} {When seeing overrides knowing: Disentangling knowledge conflicts in vision-language models}.
\newblock \emph{Preprint}, arXiv:2507.13868.

\bibitem[{Shannon(1948)}]{6773024}
C.~E. Shannon. 1948.
\newblock \href {https://doi.org/10.1002/j.1538-7305.1948.tb01338.x} {A mathematical theory of communication}.
\newblock \emph{The Bell System Technical Journal}, 27(3):379--423.

\bibitem[{Sivakumar et~al.(2025)Sivakumar, Zhang, Hakim et~al.}]{sivakumar2025steervlm}
Ashwin Sivakumar, Allen Zhang, Zaid Hakim, and 1 others. 2025.
\newblock {SteerVLM}: Robust model control through lightweight activation steering for vision language models.
\newblock In \emph{Findings of the Association for Computational Linguistics: EMNLP 2025}, pages 23640--23665.

\bibitem[{Stroop(1935)}]{stroop1935studies}
J~Ridley Stroop. 1935.
\newblock Studies of interference in serial verbal reactions.
\newblock \emph{Journal of experimental psychology}, 18(6):643.

\bibitem[{Turpin et~al.(2023)Turpin, Michael, Perez, and Bowman}]{turpin2023language}
Miles Turpin, Julian Michael, Ethan Perez, and Samuel~R. Bowman. 2023.
\newblock Language models don't always say what they think: Unfaithful explanations in chain-of-thought prompting.
\newblock In \emph{Advances in Neural Information Processing Systems}, volume~36, pages 74952--74965.

\bibitem[{Wang et~al.(2023)Wang, Xie, Jiang, Mandlekar, Xiao, Zhu, Fan, and Anandkumar}]{wang2023voyager}
Guanzhi Wang, Yuqi Xie, Yunfan Jiang, Ajay Mandlekar, Chaowei Xiao, Yuke Zhu, Linxi Fan, and Anima Anandkumar. 2023.
\newblock Voyager: An open-ended embodied agent with large language models.
\newblock \emph{arXiv preprint arXiv:2305.16291}.

\bibitem[{Wang et~al.(2026{\natexlab{a}})Wang, Li, Zhang et~al.}]{wang2026vfat}
Jingyi Wang, Ming Li, Hao Zhang, and 1 others. 2026{\natexlab{a}}.
\newblock {V-FAT}: Benchmarking visual fidelity against text-bias.
\newblock \emph{arXiv preprint arXiv:2601.04897}.

\bibitem[{Wang et~al.(2026{\natexlab{b}})Wang, Cao, Zhang, Yuan, Shan, Chen, and Gao}]{wang2024vlbiasbench}
Sibo Wang, Xiangkui Cao, Jie Zhang, Zheng Yuan, Shiguang Shan, Xilin Chen, and Wen Gao. 2026{\natexlab{b}}.
\newblock \href {https://doi.org/10.1109/TPAMI.2026.3683747} {{VLBiasBench}: A comprehensive benchmark for evaluating bias in large vision-language model}.
\newblock \emph{IEEE Transactions on Pattern Analysis and Machine Intelligence}, pages 1--14.

\bibitem[{{xAI}(2025)}]{xai2026grok}
{xAI}. 2025.
\newblock Grok 4 fast: Cost-efficient reasoning at scale.
\newblock \url{https://x.ai/news/grok-4-fast}.
\newblock Accessed: 2026-05-24.

\bibitem[{Yang et~al.(2023)Yang, Yue, and He}]{yang2023auto}
Hui Yang, Sifu Yue, and Yunzhong He. 2023.
\newblock Auto-gpt for online decision making: Benchmarks and additional opinions.
\newblock \emph{arXiv preprint arXiv:2306.02224}.

\bibitem[{Yao et~al.(2022)Yao, Zhao, Yu, Du, Shafran, Narasimhan, and Cao}]{yao2022react}
Shunyu Yao, Jeffrey Zhao, Dian Yu, Nan Du, Izhak Shafran, Karthik Narasimhan, and Yuan Cao. 2022.
\newblock React: Synergizing reasoning and acting in language models.
\newblock \emph{arXiv preprint arXiv:2210.03629}.

\bibitem[{Ye et~al.(2024)Ye, Liu, Zheng et~al.}]{ye2024mm}
Wenqian Ye, Bo~Liu, Guangtao Zheng, and 1 others. 2024.
\newblock {MM-SpuBench}: Towards better understanding of spurious biases in multimodal {LLMs}.
\newblock In \emph{Advances in Neural Information Processing Systems}, volume~37.

\bibitem[{Zhang et~al.(2025)Zhang, Wang, Gong, Shi, Wang, Wang, and Hu}]{zhang2025when}
Zhuoran Zhang, Tengyue Wang, Xilin Gong, Yang Shi, Haotian Wang, Di~Wang, and Lijie Hu. 2025.
\newblock When modalities conflict: How unimodal reasoning uncertainty governs preference dynamics in {MLLMs}.
\newblock \emph{arXiv preprint arXiv:2511.02243}.

\bibitem[{Zhao et~al.(2025)Zhao, Yang, Liang, Zhao, Guo, and Wang}]{zhao2025distributionallyrobustgraphoutofdistribution}
Chu Zhao, Enneng Yang, Yuliang Liang, Jianzhe Zhao, Guibing Guo, and Xingwei Wang. 2025.
\newblock \href {https://arxiv.org/abs/2501.15555} {Distributionally robust graph out-of-distribution recommendation via diffusion model}.
\newblock \emph{Preprint}, arXiv:2501.15555.

\bibitem[{Zhao et~al.(2024)Zhao, Wang, Liu, Cheng, Aggarwal, and Derr}]{zhao2024fairnessdiversityrecommendersystems}
Yuying Zhao, Yu~Wang, Yunchao Liu, Xueqi Cheng, Charu Aggarwal, and Tyler Derr. 2024.
\newblock \href {https://arxiv.org/abs/2307.04644} {Fairness and diversity in recommender systems: A survey}.
\newblock \emph{Preprint}, arXiv:2307.04644.

\bibitem[{Zhao et~al.(2021)Zhao, Wallace, Feng, Klein, and Singh}]{zhao2021calibrate}
Zihao Zhao, Eric Wallace, Shi Feng, Dan Klein, and Sameer Singh. 2021.
\newblock Calibrate before use: Improving few-shot performance of language models.
\newblock In \emph{Proceedings of the 38th International Conference on Machine Learning}, pages 12697--12706.

\bibitem[{Zheng et~al.(2024)Zheng, Zhou, Meng, Zhou, and Huang}]{zheng2024large}
Chujie Zheng, Hao Zhou, Fandong Meng, Jie Zhou, and Minlie Huang. 2024.
\newblock Large language models are not robust multiple choice selectors.
\newblock In \emph{Proceedings of the International Conference on Learning Representations}.

\bibitem[{Zheng and Zhang(2025)}]{zheng2025modalitybiaslvlmsanalyzing}
Haohan Zheng and Zhenguo Zhang. 2025.
\newblock \href {https://arxiv.org/abs/2508.02419} {Modality bias in lvlms: Analyzing and mitigating object hallucination via attention lens}.
\newblock \emph{Preprint}, arXiv:2508.02419.

\bibitem[{Zheng et~al.(2026)Zheng, Yuan, Yao, Gu, Zheng, and He}]{zheng2026mitigating}
Jinquan Zheng, Jia Yuan, Jiacheng Yao, Chenyang Gu, Pujun Zheng, and Guoxiu He. 2026.
\newblock Mitigating selection bias in large language models via permutation-aware {GRPO}.
\newblock In \emph{Proceedings of the 64th Annual Meeting of the Association for Computational Linguistics (Volume 1: Long Papers)}.
\newblock Accepted.

\bibitem[{Zhu et~al.(2026)Zhu, Liang, Jiang, Fu, Liu, Sun, Ng, and Qin}]{zhu2026analyzingreasoningconsistencylarge}
Zhihao Zhu, Jiafeng Liang, Shixin Jiang, Jinlan Fu, Ming Liu, Guanglu Sun, See-Kiong Ng, and Bing Qin. 2026.
\newblock \href {https://arxiv.org/abs/2601.04073} {Analyzing reasoning consistency in large multimodal models under cross-modal conflicts}.
\newblock \emph{Preprint}, arXiv:2601.04073.

\end{thebibliography}

\appendix

\section{Case Study}
\label{app:CaseStudy}

\subsection{Probabilistic Event Simulation}
\label{app:coin}
A core expectation of a random selection instruction is that models should be capable of simulating fair probabilistic events.
We examine coin toss, integer generation on $[-10,10]$, and random PIN digit generation.


\paragraph{Coin Toss}

The prompt asks the model to output ``Heads'' or ``Tails'' as a simulation of a fair coin toss ($K{=}2$).
Results in Table~\ref{tab:coin_flip_results} reveal strong language-conditioned randomness bias rather than a stable model-level behavior.

Several models exhibit near-uniform behavior in one language while collapsing in another.
For example, GPT 5.1 and Grok 4 Fast remain close to the ideal 50/50 distribution under Chinese (54/46 and 52/48, respectively), yet GPT 5.1 becomes highly skewed under French and Spanish.
Similarly, Claude Sonnet 4.6 outputs almost exclusively ``Tails'' under Chinese, English, and Spanish, but reverses toward ``Heads'' under French.

Extreme collapse is also common.
Gemini-3.1 outputs the same side in all Chinese trials, while Claude Sonnet 4.6 does the same under Chinese.
Other models such as Doubao Seed 1.6 and Kimi-K2.5 exhibit severe but non-deterministic preferences, often exceeding 90/10 splits in certain languages.

Overall, the results suggest that even minimal stochastic simulation is strongly conditioned on linguistic surface form.
Randomness failures are therefore not solely model-specific, but highly language-dependent.

\begin{table}[t]
\renewcommand\arraystretch{0.75} 
\setlength{\tabcolsep}{4.pt}
\scriptsize
\centering
\begin{tabular}{l|c|c}
\toprule
\specialrule{0em}{0.8pt}{0.8pt}
\multirow{2}{*}{\textbf{Model}} & \multirow{2}{*}{\textbf{Lang}} & \textbf{Choice Distribution} \\ 
 & & \textbf{\colorbox{wincolor}{\ \ Heads\ \ }\ \ \ /\ \ \ \colorbox{losscolor}{\ \ Tails\ \ }} \\ 
\midrule
\multirow{4}{*}[-2pt]{\textsc{GPT 5.1}}               
 & EN & \hbartwo{84.0}{16.0} \\
 & ZH & \hbartwo{54.0}{46.0} \\
 & FR & \hbartwo{96.0}{4.0} \\
 & ES & \hbartwo{2.0}{98.0} \\
\midrule
\multirow{4}{*}[-2pt]{\begin{tabular}[c]{@{}l@{}}\textsc{Gemini-3.1-}\\[4pt] \textsc{flash-lite}\end{tabular}} 
 & EN & \hbartwo{70.0}{30.0} \\
 & ZH & \hbartwo{100.0}{0.0} \\
 & FR & \hbartwo{60.0}{40.0} \\
 & ES & \hbartwo{50.0}{50.0} \\
\midrule
\multirow{4}{*}[-2pt]{\begin{tabular}[c]{@{}l@{}}\textsc{Claude-}\\[4pt] \textsc{Sonnet-4.6}\end{tabular}}     
 & EN & \hbartwo{0.0}{100.0} \\
 & ZH & \hbartwo{4.0}{96.0} \\
 & FR & \hbartwo{90.0}{10.0} \\
 & ES & \hbartwo{2.0}{98.0} \\
\midrule
\multirow{4}{*}[-2pt]{\begin{tabular}[c]{@{}l@{}}\textsc{Doubao-}\\[4pt] \textsc{Seed-1.6}\end{tabular}} 
 & EN & \hbartwo{94.0}{6.0} \\
 & ZH & \hbartwo{96.0}{4.0} \\
 & FR & \hbartwo{70.0}{30.0} \\
 & ES & \hbartwo{78.0}{22.0} \\
\midrule
\multirow{4}{*}[-2pt]{\textsc{Grok 4 Fast}}           
 & EN & \hbartwo{34.0}{66.0} \\
 & ZH & \hbartwo{52.0}{48.0} \\
 & FR & \hbartwo{64.0}{36.0} \\
 & ES & \hbartwo{72.0}{28.0} \\
\midrule
\multirow{4}{*}[-2pt]{\textsc{Kimi-k2.5}}             
 & EN & \hbartwo{36.0}{64.0} \\
 & ZH & \hbartwo{98.0}{2.0} \\
 & FR & \hbartwo{66.0}{34.0} \\
 & ES & \hbartwo{22.0}{78.0} \\
\midrule
\multirow{4}{*}[-2pt]{\begin{tabular}[c]{@{}l@{}}\textsc{Qwen-}\\[4pt] \textsc{3.6-plus}\end{tabular}}          
 & EN & \hbartwo{32.0}{68.0} \\
 & ZH & \hbartwo{74.0}{26.0} \\
 & FR & \hbartwo{46.0}{54.0} \\
 & ES & \hbartwo{28.0}{72.0} \\
\specialrule{0em}{0.8pt}{0.8pt}
\bottomrule
\end{tabular}
\caption{Choice distribution of LLMs in coin flipping.}
\label{tab:coin_flip_results}
\end{table}

\paragraph{Random Integer on $[-10,10]$}

The model is instructed to generate a uniformly random integer in $[-10,10]$ ($K{=}21$).
\textbf{Claude‑4.6} collapses completely in both languages (RI${=}0$), always outputting the same integer.
\textbf{GPT‑5.1} and \textbf{Grok‑4‑fast} maintain moderate randomness (RI between $0.57$ and $0.87$), while \textbf{Doubao} and \textbf{Qwen} achieve the highest entropy (RI${>}0.80$ in English, $0.63$--$0.86$ in Chinese).
\textbf{Kimi‑k2.5} shows a striking language effect: nearly uniform under English (RI${=}0.73$) but collapses completely under Chinese (RI${=}0$).
Notably, even among models that retain some stochasticity, the empirical distributions are far from uniform: most models exhibit systematic preferences for $\pm3$ and $\pm7$, with these values receiving disproportionately high probability mass---an effect most pronounced in the models with the lowest RI scores.

\paragraph{Random Four‑Digit PIN}

The model generates a random PIN from $0000$--$9999$ ($K{=}10000$, the largest option space in the benchmark).
Despite the vast output space, \textbf{Claude‑4.6} still collapses in both languages (RI${<}0.27$), repeatedly generating the same PIN.
Most other models achieve near‑perfect uniformity (RI${>}0.87$), with \textbf{Qwen‑3.6} reaching RI${=}0.99$ under both languages, demonstrating that large option spaces can partially mitigate stochastic collapse for some models.

\subsection{Cross-Modal Hijacking}
\label{app:Hijacking}

We examine two Stroop-like conflicts designed to probe modality prioritization under logic‑neutral instructions. A milder label–image mismatch case (Case~\#42229416 vs.\ Case~\#ee5542eb, fruit images with partially swapped labels) produced comparatively modest RI shifts, yet still confirmed that textual interference can reroute a model's implicit preference without restoring genuine uniformity.

\subsubsection{Case~\#2d27488e: The Cat-``DOG'' Paradigm}

\textbf{Setup.} A simple line drawing of a cat with the word ``DOG'' superimposed. The instruction is ``Name the animal you see'' ($K=2$). We note that the cat image is a line drawing with relatively low visual feature density; more photorealistic stimuli might shift the cross‑modal weighting of some models, although the current paradigm already reveals a strong text‑dominant prior.

\textbf{Results.} In both English and Chinese, six of seven models output ``Dog'' in all 50 trials ($\text{RI}=0.00$). \textbf{Grok 4 Fast} under English selects ``Cat'' 12 times and ``Dog'' 38 times ($\text{RI}=0.80$), but collapses entirely to ``Dog'' in Chinese.

\textbf{Key findings.} This item achieves near‑universal collapse---the textual label overrides the visual content in virtually all model–language combinations, providing the strongest evidence for cross‑modal hijacking in the entire benchmark.

\subsubsection{Case~\#5665b6a2: Explicit Instruction vs.\ Randomness Requirement}

\textbf{Setup.} The prompt displays ``Please be sure to select B'' alongside four buttons A--D. Models are instructed to ``randomly select a button'' ($K=4$).

\textbf{Results.} Most models largely comply with the explicit directive, selecting B in the majority of trials (\textbf{Gemini-3.1}, \textbf{Qwen-3.6} and \textbf{Grok 4 Fast} under Chinese select B in all 50 trials). \textbf{Kimi-k2.5} is the striking outlier: under English it deliberately avoids B (spreading selections across A, C, D; $\text{RI}=0.34$), but under Chinese it collapses entirely onto A (50/50). Notably, \textbf{Kimi-k2.5} spontaneously justified its English choice on two trials, explicitly stating that the ``random'' instruction overrides the image directive (e.g., ``\textit{you asked for a random choice, so I selected A instead}''). Despite this conflict awareness, the model merely shifted its deterministic lock‑in from B to another option rather than achieving uniform randomness.

\subsection{Visual Saliency Preference}
\label{app:visual_saliency}

We examine whether subtle visual cues---imperceptible or barely noticeable to humans---can bias a model's random selection despite explicit instructions to act randomly.
The five items analysed in this category are Hashes Case~\#0327dda4, Case~\#3edeab77, Case~\#d02cf130, Case~\#da34295d, and Case~\#a33f51c3.
All items present fine-grained visual differences that are irrelevant to the random selection task, testing whether bottom‑up perceptual saliency can override top‑down randomness intentions.

\subsubsection{Case~\#da34295d: 1\% Gaussian Blur}

\textbf{Setup.}
Four identical roses are arranged horizontally, one of which has a 1\% Gaussian blur applied to its edge ($K=4$).
The blur is below the just‑noticeable‑difference threshold for casual human viewing and was independently verified to be imperceptible without deliberate comparison.

\textbf{Choice distributions.}
The dominant selections (counts $\ge 10$ out of 50 trials) are reported below; P3 denotes the blurred rose.

\begin{table}[h]
\centering
\begin{tabular}{lrl}
\toprule
\textbf{Model} & \textbf{Lang} & \textbf{Dominant choices}\\
\midrule
\multirow{2}{*}{\textbf{Claude Sonnet 4.6}}  & EN & P3: 44 \\
                              & ZH & P3: 50 \\[2pt]
\multirow{2}{*}{\textbf{Gemini-3.1-flash}}  & EN & P3: 30, P2: 18 \\
                              & ZH & P3: 32, P2: 15 \\[2pt]
\multirow{2}{*}{\textbf{Grok 4 Fast}} & EN & P3: 36 \\
                              & ZH & P3: 25, P2: 21 \\[2pt]
\multirow{2}{*}{\textbf{GPT 5.1}}     & EN & P3: 25, P1: 15 \\
                              & ZH & P4: 27, P2: 13 \\[2pt]
\multirow{2}{*}{\textbf{Qwen 3.6 Plus}}    & EN & P3: 28, P2: 22 \\
                              & ZH & P3: 25, P2: 21 \\[2pt]
\multirow{2}{*}{\textbf{Kimi-k2.5}}   & EN & P2: 37, P3: 12 \\
                              & ZH & P3: 26, P2: 24 \\[2pt]
\multirow{2}{*}{\textbf{Doubao Seed 1.6}}      & EN & P2: 41 \\
                              & ZH & P2: 29, P1: 15 \\
\bottomrule
\end{tabular}

\caption{Dominant choices for the 1\% Gaussian Blur item (Case~\#da34295d).}
\label{tab:blur_choices}
\end{table}

\textbf{Sensitivity tiers.}
The models exhibit three distinct levels of response to the blur.
\textbf{Claude-4.6} is maximally sensitive: it locks onto P3 almost exclusively (44/50 EN, 50/50 ZH), showing that the blur fully overrides the randomness instruction.
\textbf{Gemini-3.1} and \textbf{Grok 4 Fast} (English) are moderately sensitive, concentrating on P3 with weaker dominance (30--36 selections) and a visible secondary mode, indicating detection of the anomaly without complete stochastic collapse.
\textbf{GPT 5.1}, \textbf{Qwen-3.6}, and \textbf{Grok 4 Fast} (Chinese) are weakly sensitive: selections are diffuse across multiple positions with no single option exceeding 30 counts.
\textbf{Kimi-k2.5} and \textbf{Doubao} deviate from the consensus by favouring P2 over P3, demonstrating that the direction of the bias is model‑specific rather than uniformly attracted to the blurred option.

\textbf{Key findings.}
A 1\% Gaussian blur---below the human just‑noticeable‑difference---induces non‑uniform choice distributions in all seven models.
The effect is graded rather than binary, ranging from complete collapse (Claude) to mild preference or divergent orientation (Kimi, Doubao).
This confirms that visual saliency can act as a bottom‑up anchor that overrides top‑down randomness instructions, but the strength and direction of this anchoring are model‑specific properties.

\subsubsection{Unifying Observation across Case~\#0327dda4, Case~\#3edeab77, Case~\#d02cf130, Case~\#da34295d, Case~\#a33f51c3}

Across the five visual saliency items, a consistent pattern emerges: \textbf{every model exhibits systematic non‑uniformity on at least one item, but no model collapses on all five.}
\textbf{Claude-4.6} is the most sensitive overall, yet its RI on Hash Case~\#da34295d EN (0.26) is far higher than its near‑zero RI on spatial or emotional dimensions.
\textbf{GPT 5.1} is the most robust on the blur item (no position exceeds 27/50), yet collapses on the brightness asymmetry of Hash Case~\#d02cf130 (RI$\approx$0 in both languages).
\textbf{Doubao} and \textbf{Grok 4 Fast} are the least affected, showing moderate‑to‑high randomness across most saliency items.

The unifying conclusion is that visual saliency does not operate as a monolithic bias: different visual features (blur, brightness, opacity, colour depth) activate distinct, model‑specific heuristic pathways.
A model that collapses on blur may remain robust to brightness, and vice versa.
This challenges the notion of a single ``visual sensitivity'' trait and instead points to a fragmented landscape of feature‑specific anchoring.

\subsection{Classic Visual Illusions}

We examine two classic perceptual illusions to investigate whether MLLMs exhibit human-like systematic preferences under ambiguous visual stimuli.

\subsubsection{Case~\#eeb1b93f: The Yin-Jiapeng (Checker-Shadow) Illusion}

\textbf{Setup.} The classic checker-shadow illusion, where squares A and B appear perceptually different in lightness due to surrounding shadow context but are objectively identical in color. The model is instructed to ``randomly pick one'' between the two squares ($K=2$).

\textbf{Results.} Model behavior on this illusion is highly heterogeneous (Table~\ref{tab:yinji}).
\textbf{Gemini-3.1} collapses completely in both languages ($\text{RI}\approx0$), selecting the same square in all 50 trials, while \textbf{Grok 4 Fast} and \textbf{Doubao} maintain relatively high randomness ($\text{RI}>0.7$). \textbf{Claude-4.6} and \textbf{GPT 5.1} exhibit RI values ranging from $0.24$ to $0.98$.

\begin{table}[h]
\centering
\caption{RI values for the Yin-Jiapeng illusion (Case~\#eeb1b93f).}
\label{tab:yinji}
\begin{tabular}{lcc}
\toprule
\textbf{Model} & \textbf{RI (EN)} & \textbf{RI (ZH)} \\
\midrule
\textbf{Claude Sonnet 4.6} & 0.00 & 0.98 \\
\textbf{Doubao Seed 1.6}     & 0.72 & 0.83 \\
\textbf{GPT 5.1}    & 0.24 & 0.94 \\
\textbf{Gemini-3.1-flash} & 0.00 & 0.00 \\
\textbf{Grok 4 Fast}& 0.88 & 0.98 \\
\textbf{Kimi-k2.5}  & 0.58 & 0.53 \\
\textbf{Qwen 3.6 Plus}   & 0.58 & 0.98 \\
\bottomrule
\end{tabular}
\end{table}

\textbf{Qualitative observation}: \textbf{Doubao} and \textbf{Kimi-k2.5} occasionally appended explanatory text to their choices, explicitly identifying the figure as the classic checker-shadow illusion and noting that the two squares are actually the same color. Nevertheless, their selection distributions were not uniform---\textbf{Doubao}, for instance, showed a systematic preference for one square (approx.\ 35/50 trials) despite demonstrating accurate declarative knowledge of the illusion. This provides direct evidence for the dual process dissociation: System~2 possesses correct explicit knowledge, but System~1 perceptual bias continues to systematically influence the final choice.

\subsubsection{Case~\#ae858218: The Ebbinghaus (Titchener Circles) Illusion}

\textbf{Setup.} The classic Ebbinghaus illusion, where two central orange dots are surrounded by gray circles of different sizes, causing humans to perceive one dot as larger. The model is instructed to ``randomly select an orange dot'' ($K=2$).

\textbf{Results.} Collapse is more widespread on this illusion (Table~\ref{tab:ebbinghaus}). Most models exhibit moderate-to-strong preferences; only \textbf{Grok 4 Fast} maintains consistently high randomness ($\text{RI}>0.94$). \textbf{Claude-4.6} and \textbf{GPT 5.1} show RI values below $0.24$ in most conditions.

\begin{table}[h]
\centering
\caption{RI values for the Ebbinghaus illusion (Case~\#ae858218).}
\label{tab:ebbinghaus}
\begin{tabular}{lcc}
\toprule
\textbf{Model} & \textbf{RI (EN)} & \textbf{RI (ZH)} \\
\midrule
\textbf{Claude Sonnet 4.6} & 0.00 & 0.92 \\
\textbf{Doubao Seed 1.6}     & 0.33 & 0.16 \\
\textbf{GPT 5.1}    & 0.14 & 0.00 \\
\textbf{Gemini-3.1-flash} & 0.63 & 0.00 \\
\textbf{Grok 4 Fast}& 0.96 & 0.93 \\
\textbf{Kimi-k2.5}  & 0.98 & 0.00 \\
\textbf{Qwen 3.6 Plus}   & 0.90 & 0.48 \\
\bottomrule
\end{tabular}
\end{table}

\textbf{Qualitative observation: A ``reverse illusion'' from \textbf{Kimi-k2.5}}: In a small number of English trials, \textbf{Kimi-k2.5} accompanied its choice with a justification, stating that the orange dot surrounded by \textit{larger} gray circles appeared larger and was therefore selected. This judgment runs \textbf{opposite} to the standard human percept, where the dot surrounded by \textit{smaller} circles is typically perceived as larger. This ``reverse illusion'' provides strong qualitative evidence that the model's preference is genuinely driven by the surrounding visual context rather than by position or random guessing, even though the resulting perceptual judgment differs from that of humans.

\subsubsection{Synthesis}

The two illusions jointly demonstrate that MLLMs are not immune to perceptual illusions. The Yin-Jiapeng results show that explicit knowledge of the illusion (\textbf{Doubao}, \textbf{Kimi}) coexists with implicit, non-uniform selection behavior, reflecting a decoupling of declarative knowledge from stochastic control. The Ebbinghaus results show that the visual context actively shapes model preferences, as evidenced by Kimi's context-driven (if human-divergent) justification.

We acknowledge a limitation of the current design: both illusions present options with inherent positional differences (left/right), and we cannot fully separate illusion effects from position bias without additional controls (e.g., rotated or mirrored variants). This remains a direction for future work.

\subsection{Positional Preference}

We examine a suite of spatial layout tasks designed to isolate position‑driven selection heuristics:
vertical drawers (Case~\#b4074266), quadrant dots (Case~\#32e3ad84), cross‑arranged squares (Case~\#25ddb936), identical paper slips (Case~\#e8356992), a $3{\times}3$ grid with a single filled cell (Case~\#aa130e4f), two balloons (Case~\#06c11919), and three arrow‑direction variants with different visual layouts (Case~\#8e290501, Case~\#6fe9bcab, Case~\#f78a9f5b).

\subsubsection{Case~\#aa130e4f: The $3\times3$ Grid}

A $3\times3$ grid contains one black cell at the top‑left ($K = 9$).
Across all models and both languages, the \textbf{center region} consistently attracts the highest probability mass, even though the explicit visual marker is the filled top‑left cell.
This indicates that endogenous spatial priors dominate an explicit task‑designed anchor.

The strength of this center‑directed collapse is language‑modulated in several models:
\textbf{Claude} under English concentrates $49/50$ selections on the middle‑right cell ($\text{RI}{=}0.045$), whereas under Chinese $35/50$ selections shift to the top‑left ($\text{RI}{=}0.447$).
\textbf{GPT‑5.1} under Chinese locks almost entirely onto the middle‑center cell ($47/50$, $\text{RI}{=}0.121$), while under English its probability mass concentrates on the bottom row ($\text{RI}{=}0.388$).
\textbf{Grok} under Chinese collapses to the top‑left ($48/50$, $\text{RI}{=}0.089$); under English its distribution is comparatively diffuse ($\text{RI}{=}0.888$), yet the center remains the most chosen cell.
These extremes illustrate how language can gate positional bias, but the underlying preference for the center is shared.

\subsubsection{Case~\#8e290501, Case~\#6fe9bcab, Case~\#f78a9f5b: Arrow Layout Variants}

Three isomorphic questions ask the model to ``randomly choose an arrow direction'', but the arrows are presented in a central cross, in four quadrants, or as a horizontal row ($K{=}4$).
All models exhibit a systematic preference for \textbf{right‑pointing and up‑pointing arrows} across all three layouts and both languages.
Models such as \textbf{Doubao}, \textbf{Grok}, and \textbf{Kimi} sometimes display a noticeable left‑arrow preference that can even surpass right‑arrow selections in a given condition, while \textbf{Kimi} is the only model that has produced a majority of down‑arrow selections in one instance.
Nevertheless, the right/up tendency remains the dominant pattern across the full set of observations.
Layout changes can modulate which of these two directions is expressed, but this modulatory effect is weaker than the underlying directional prior.

\subsubsection{Unifying Observations}

Across all positional tasks, several consistent patterns emerge.
\textbf{Center, right, and top constitute the most prevalent preference directions}: middle‑right and center in the grid (Case~\#aa130e4f), upper drawer in Case~\#b4074266, center square in Case~\#25ddb936, right/up arrows in the three arrow tasks, and the top balloon in Case~\#06c11919.
Even when all visual content is removed (identical paper slips, Case~\#e8356992), several models still lock onto a specific ordinal position.
\textbf{Positional preference is a robust, cross‑task determinant of model choice under logic‑neutral instructions}, persisting across varied spatial layouts and often exceeding the influence of explicit visual markers.

\subsection{Other Key Observations}

\textbf{Emotional Preference.}
Across both photo (Case~\#5bec4d6d, Case~\#0556cea1, Case~\#637048b6, Case~\#51f5b6e7, Case~\#d82ce04d, Case~\#a4c3fb01, Case~\#c020017f, Case~\#86ac7f4b) and text (Case~\#9c1fe4d0, Case~\#55f74ec9, Case~\#2bf2ca21, Case~\#855144fb) emotion‑laden tasks, most models exhibit systematic biases, although no universal preference direction emerges.
Some models consistently favor positive emotions, others lean toward negative or neutral ones, and \textbf{Grok‑4‑fast} shows notably lower overall emotional anchoring.
The heterogeneity suggests that emotional valence is encoded in model‑specific ways rather than reflecting a shared human‑like positivity bias.

\textbf{Social Identity.}
Several social‑identity tasks produce near‑universal collapse.
On Case~\#e62a2d67 (``90\% success rate'' vs.\ ``10\% failure rate''), all models except \textbf{Qwen‑3.6} under English overwhelmingly select the positively framed option, making this one of the most severe collapse sites in the benchmark.
Case~\#a0fc0e1a (option popularity) reveals strong conformity effects: most models gravitate toward the option with the highest displayed count, although a non‑negligible fraction instead favors the least popular alternative.
Case~\#b2b7a525 (default selection) exerts a weaker but still observable anchoring influence.
Case~\#ad28a7c4 (``Recommended'' label) induces overwhelming convergence on the recommended item for every model.
Case~\#06a0a56f (scarcity: ``Only 1 left'') elicits strong preference from the majority of models, while Case~\#1344817f (price tags) shows markedly lower value‑based bias across all models.
Case~\#3d0c3753 and Case~\#c7bde32f (profession, age, gender) yield model‑dependent patterns, with firefighters, elderly figures, and young girls receiving consistently elevated selection probabilities.
Case~\#eef70b6e (``Yes'' / ``No'' / ``Maybe'') reveals a cross‑model preference for ``Maybe'', with some models additionally favoring ``No'' in certain languages.

\textbf{Complexity and Order.}
Preferences between orderly and chaotic visual stimuli (Case~\#46e2933a, Case~\#8df3da62, Case~\#047e7eb1, Case~\#22ccf1bb, Case~\#266e807d, Case~\#293374a5, Case~\#e7e31114, Case~\#00533ec7, Case~\#8f2bf139) are highly model‑specific.
\textbf{Gemini‑3.1} consistently favors clean, regular patterns, while \textbf{Claude‑4.6} shows a mild preference for slight irregularity.
Overall randomness on these tasks remains relatively high compared to other bias categories, indicating that complexity preference is a weaker and more idiosyncratic anchoring dimension.

\textbf{Zero‑Information Selection.}
Text Case~\#07f08546 presents four completely empty options, removing all content‑based cues.
This task should, in principle, yield the most uniform distribution in the benchmark.
Instead, substantial bias is observed: the majority of models exhibit a systematic preference for option~C, confirming that the option label itself---independent of any semantic content---can anchor model decisions.

\textbf{Semantic Equivalence.}
Across 45 text tasks with strictly equivalent options, \textbf{Claude‑4.6} under English and \textbf{Kimi‑k2.5} under Chinese systematically default to positional or lexicographic heuristics, while other models exhibit varying degrees of robustness.

\section{Additional Details For Dataset Curation}

\begin{figure*}[!t]
\begin{tcolorbox}[colback=myblack, colframe=myframe, title=\textbf{Example for Image Generation Prompt.}]
\begin{verbatim}
A meticulously aligned four-panel visual stimulus configured into a strict 2x2 
grid matrix, tracking the exact same geographic vantage point across four se-
quential seasonal states. 

The underlying geometry, composition, and camera perspective must remain rigidly 
locked and perfectly synchronized across all four panels: a wide-angle landscape 
shot featuring a centralized wood-plank boardwalk that extends symmetrically from 
the bottom-center foreground into a tranquil body of water, leading toward a 
clean horizon line with soft mountain silhouettes. 

The landscape transitions fluidly across the quadrants, tracking a soft morning 
spring scene with flowers blossoms in full bloom and fresh green banks (top-left), 
a vivid summer state with dense emerald trees under a radiant midday sun (top-
right), a misty autumn after-noon with fallen golden-yellow and auburn leaves 
carpeting the wooden path (bottom-left), and a frozen winter twilight blanketed 
in thick white snow and frost over the dormant landscape (bottom-right). 

Style Constraints: Professional award-winning landscape photography, ultra-sharp 
detail, high contrast, clean minimalist white lines separating the 4 quadrants 
into an exact symmetrical grid. Absolute zero text, zero watermarks, and zero arti-
fact distortions.
\end{verbatim}
\end{tcolorbox}
\caption{Image Generation Prompt.}
\label{fig:prompt}
\end{figure*}

\paragraph{Vision Stimulus Generation Prompt}
\label{app:prompt}
To evaluate the model's inductive bias over temporal-seasonal transitions and spatial grid positions without introducing visual confounding variables, we design a highly structured prompt for generative vision APIs. The exact prompt used is formalized in Figure~\ref{fig:prompt}.

\section{Additional Details For Main Results}
\label{app:heatmap}

\begin{figure*}[!t]
\centering
\begin{subfigure}[b]{\textwidth}
   \centering
   \includegraphics[width=\textwidth]{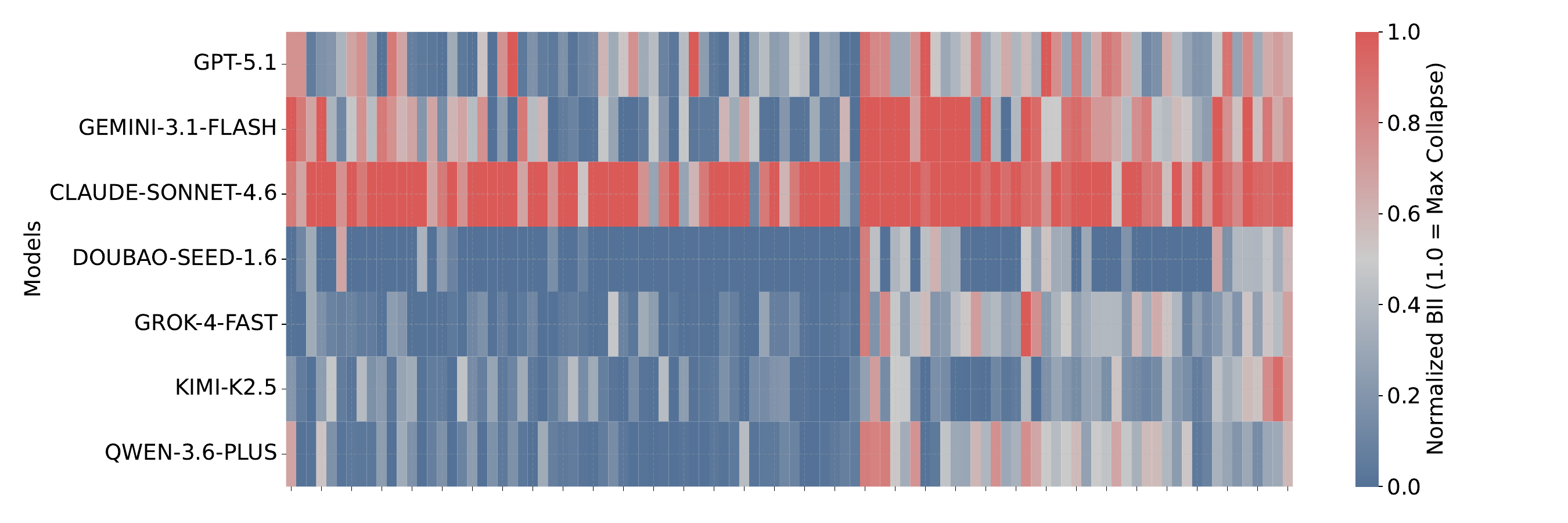}
   \caption{RB-Text Modality}
   \label{fig:heatmap_text_en}
\end{subfigure}
\hfill
\begin{subfigure}[b]{\textwidth}
   \centering
   \includegraphics[width=\textwidth]{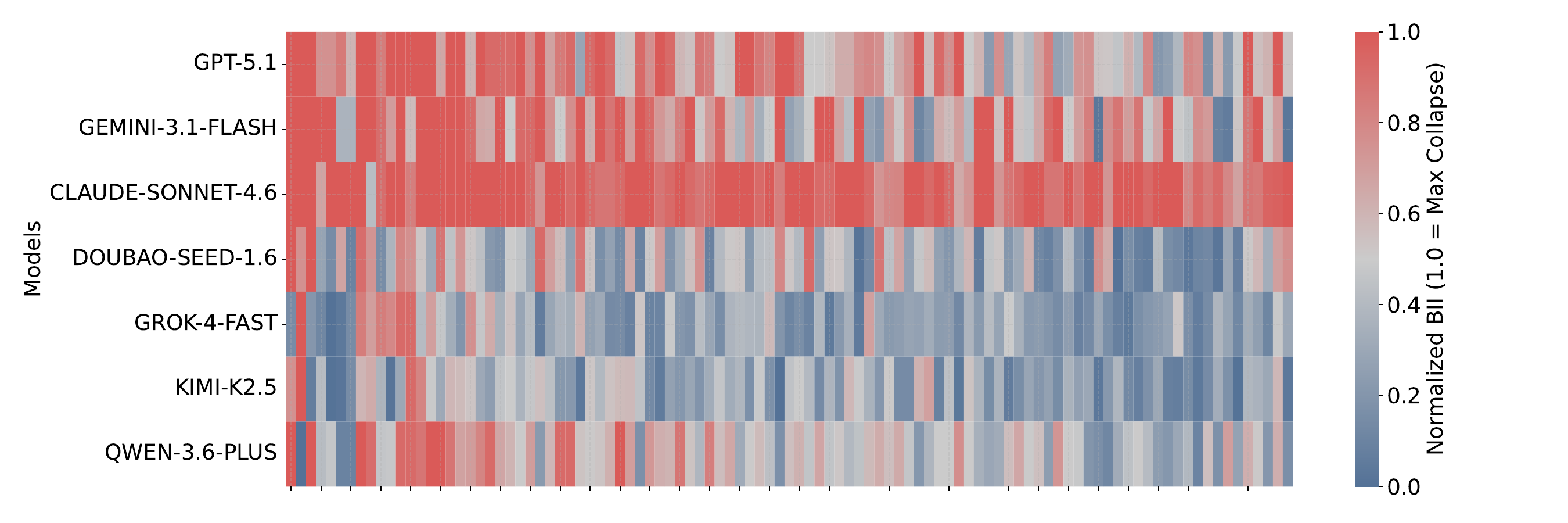}
   \caption{RB-Vision Modality}
   \label{fig:heatmap_photo_en}
\end{subfigure}
\caption{Cross-Modal Comparison of Bias Intensity ($BII$) Heatmaps.}
\label{fig:modality_heatmaps_comparison}
\end{figure*}

\begin{figure*}[!t]
\centering
\begin{subfigure}[b]{0.45\textwidth}
   \centering
   \includegraphics[width=\textwidth]{fig/radar_EN.pdf}
   \caption{Radar Chart of English $RI$ Results}
   \label{fig:rader_en}
\end{subfigure}
\hfill
\begin{subfigure}[b]{0.45\textwidth}
   \centering
   \includegraphics[width=\textwidth]{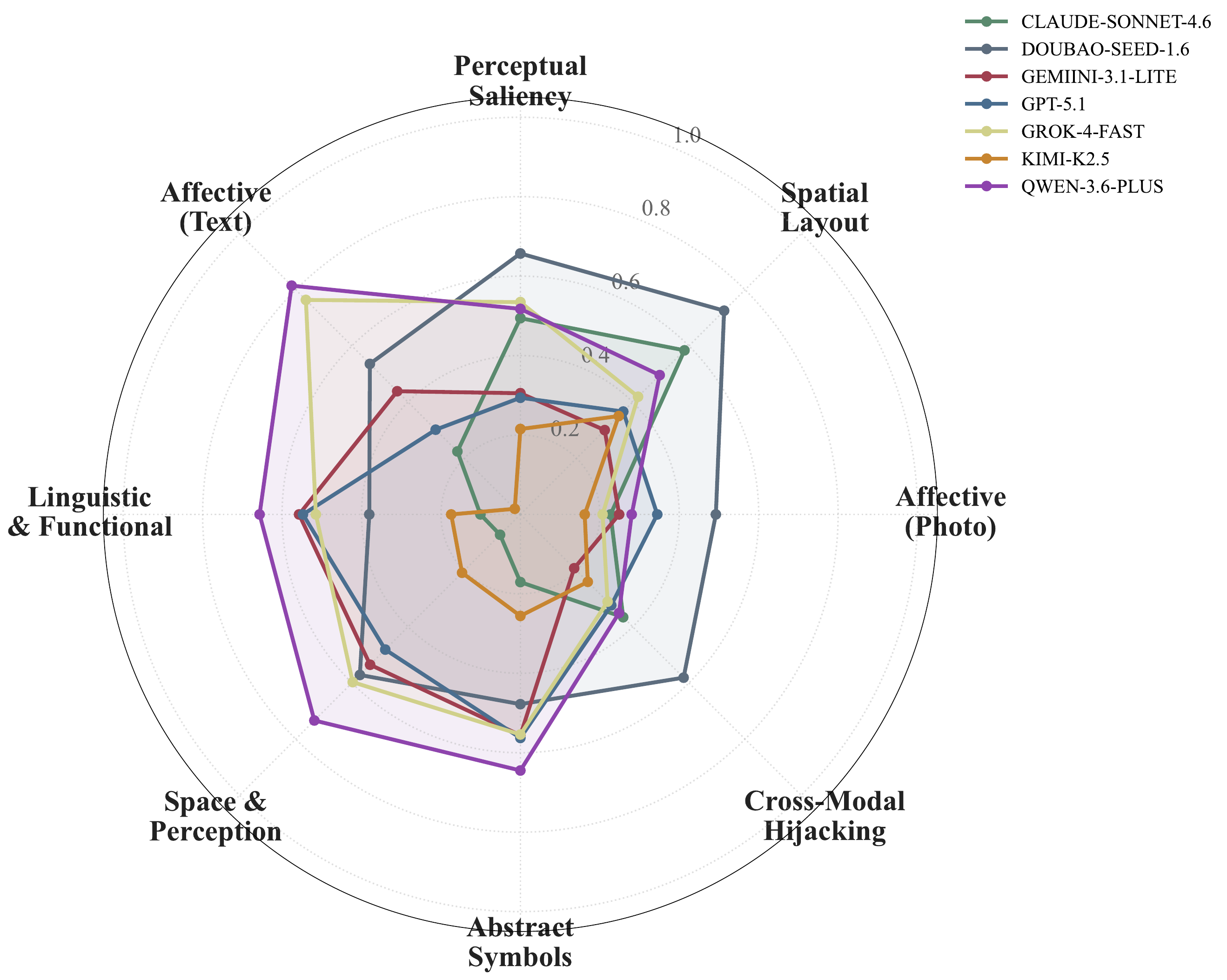}
   \caption{Radar Chart of Chinese $RI$ Results}
   \label{fig:rader_zh}
\end{subfigure}
\caption{Radar chart of $RI$ score across cognitive dimensions and subcategories.}
\label{fig:rader}
\end{figure*}

Figure~\ref{fig:modality_heatmaps_comparison}  shows the empirical Bias Intensity ($BII$) of MLLMs between the RB-Text (\textbf{a}) and RB-Vision (\textbf{b}) in English.
The results show that the introduction of visual modalities leads to a substantial reduction in stochasticity across all evaluated MLLMs.

Figure~\ref{fig:rader} presents the performance of different models across fine-grained categories under both English and Chinese prompts.

\section{Additional Details For Ablation Analysis}
\subsection{Metric}
\label{app:jsd_analysis}

To mathematically quantify the perturbation of the models' implicit decision manifolds caused by distinct ablation variables (e.g., linguistic context switching and option symbol substitution), we introduce the \textbf{Jensen-Shannon Divergence (JSD)} into our detailed ablation analysis.

Unlike $BII$, which measures the distance between an empirical distribution and an absolute uniform baseline, $JSD$ serves as a symmetric and smoothed divergence metric. It evaluates the topological discrepancy between two empirical probability distributions, $P_{V_1}$ and $P_{V_2}$, generated by the identical model under the same logic-neutral task but governed by different control variables (e.g., $V_1 = \text{English}$, $V_2 = \text{Chinese}$):

\begin{align}
&JSD(P_{V_1} \parallel P_{V_2}) \nonumber \\
& = \frac{1}{2} D_{KL}(P_{V_1} \parallel M) + \frac{1}{2} D_{KL}(P_{V_2} \parallel M)
\end{align}

where $M = \frac{1}{2}(P_{V_1} + P_{V_2})$ represents the mixture distribution. The resulting $JSD$ is bounded within $[0, 1]$ (when utilizing base-2 logarithms). A low $JSD$ indicates that the model's implicit preferences are robust against the manipulated variable. Conversely, a high $JSD$ provides definitive empirical evidence that the control variable (such as linguistic syntax or token representation) acts as an orthogonal routing switch, directly triggering fundamentally distinct heuristic decision pathways within the parameter space.

\subsection{Cross-Lingual Analysis}
\label{app:ablation_language}
This section provides a granular quantitative dissection of the heterogeneous mode collapse exhibited by frontier models when subjected to identical visual stimuli but different linguistic instructions.

\begin{figure*}[!t]
    \centering
    \includegraphics[width=\textwidth]{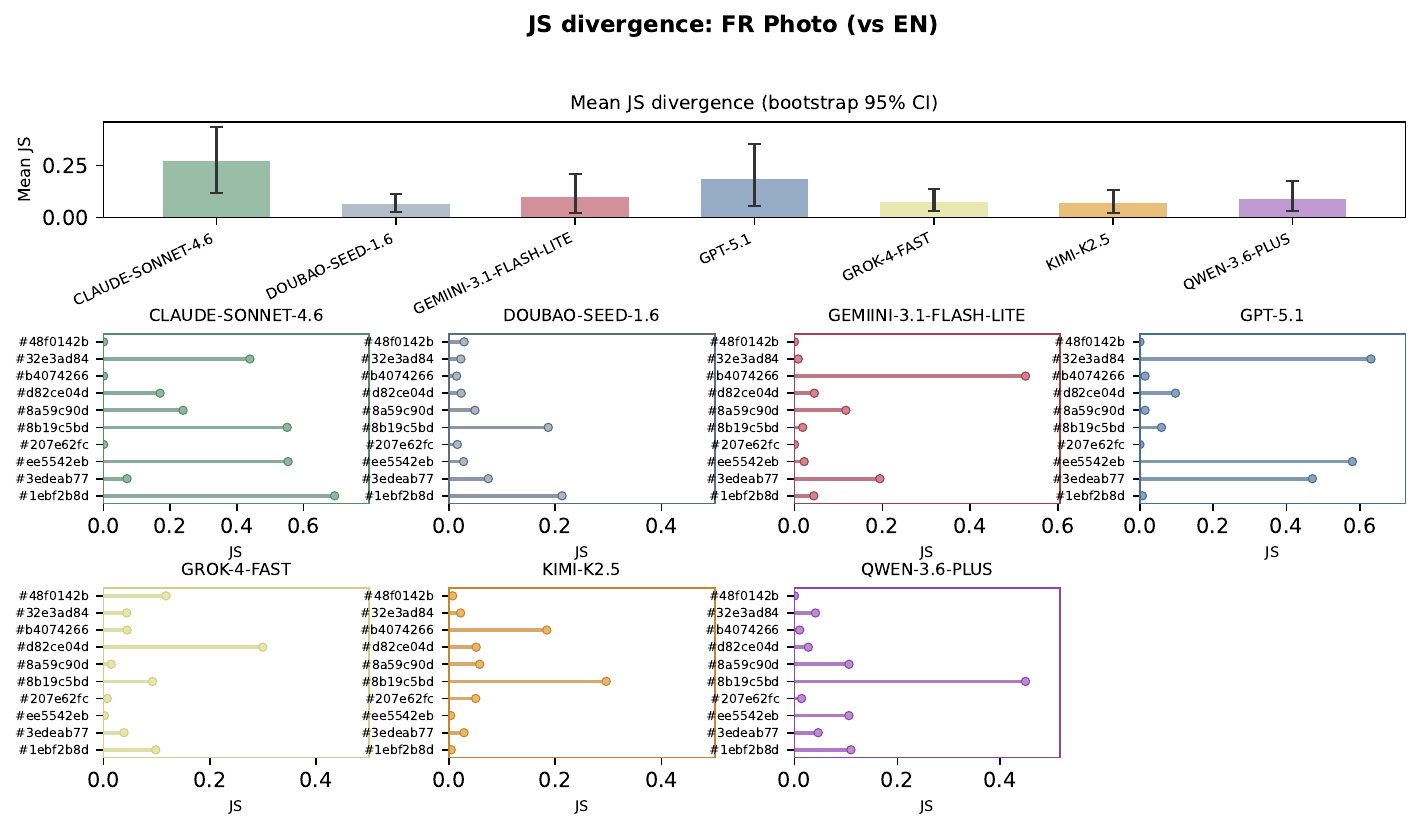}
    \caption{JS divergence: FR Photo vs. EN}
    \label{fig:js_fr_photo}
\end{figure*}

\begin{figure*}[!t]
    \centering
    \includegraphics[width=\textwidth]{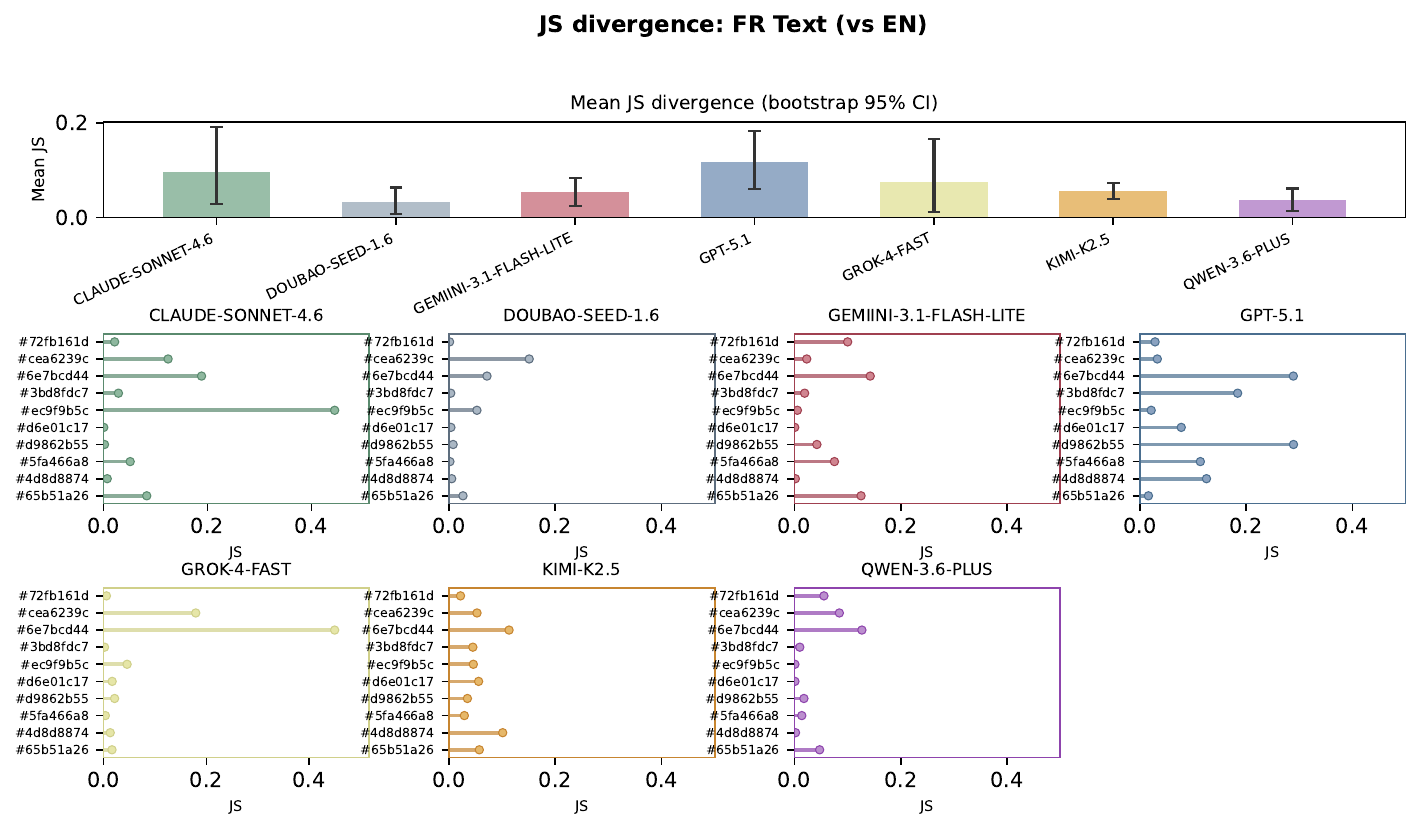}
    \caption{JS divergence: FR Text vs. EN}
    \label{fig:js_fr_text}
\end{figure*}

\begin{figure*}[!t]
    \centering
    \includegraphics[width=\textwidth]{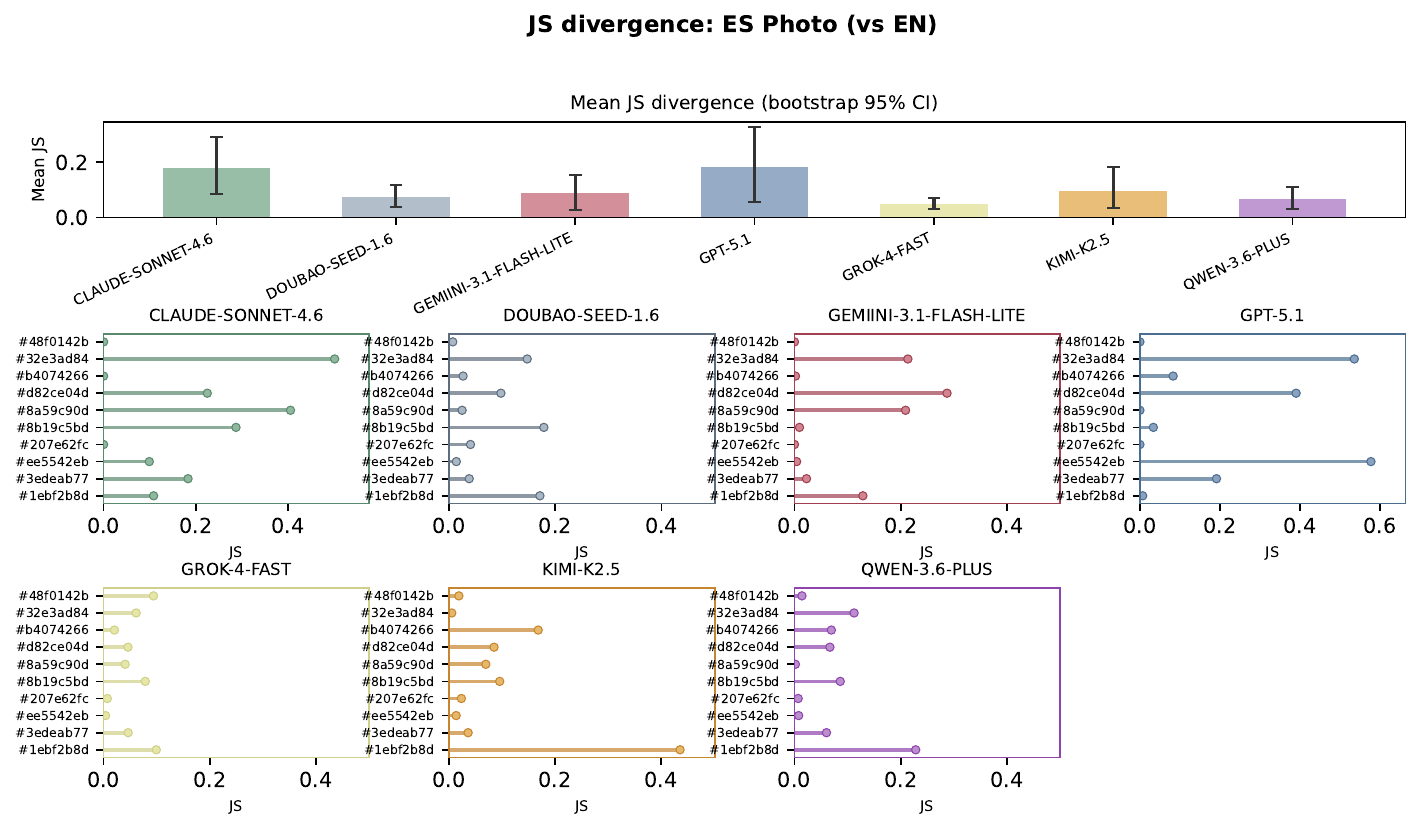}
    \caption{JS divergence: ES Photo vs. EN}
    \label{fig:js_es_photo}
\end{figure*}

\begin{figure*}[!t]
    \centering
    \includegraphics[width=\textwidth]{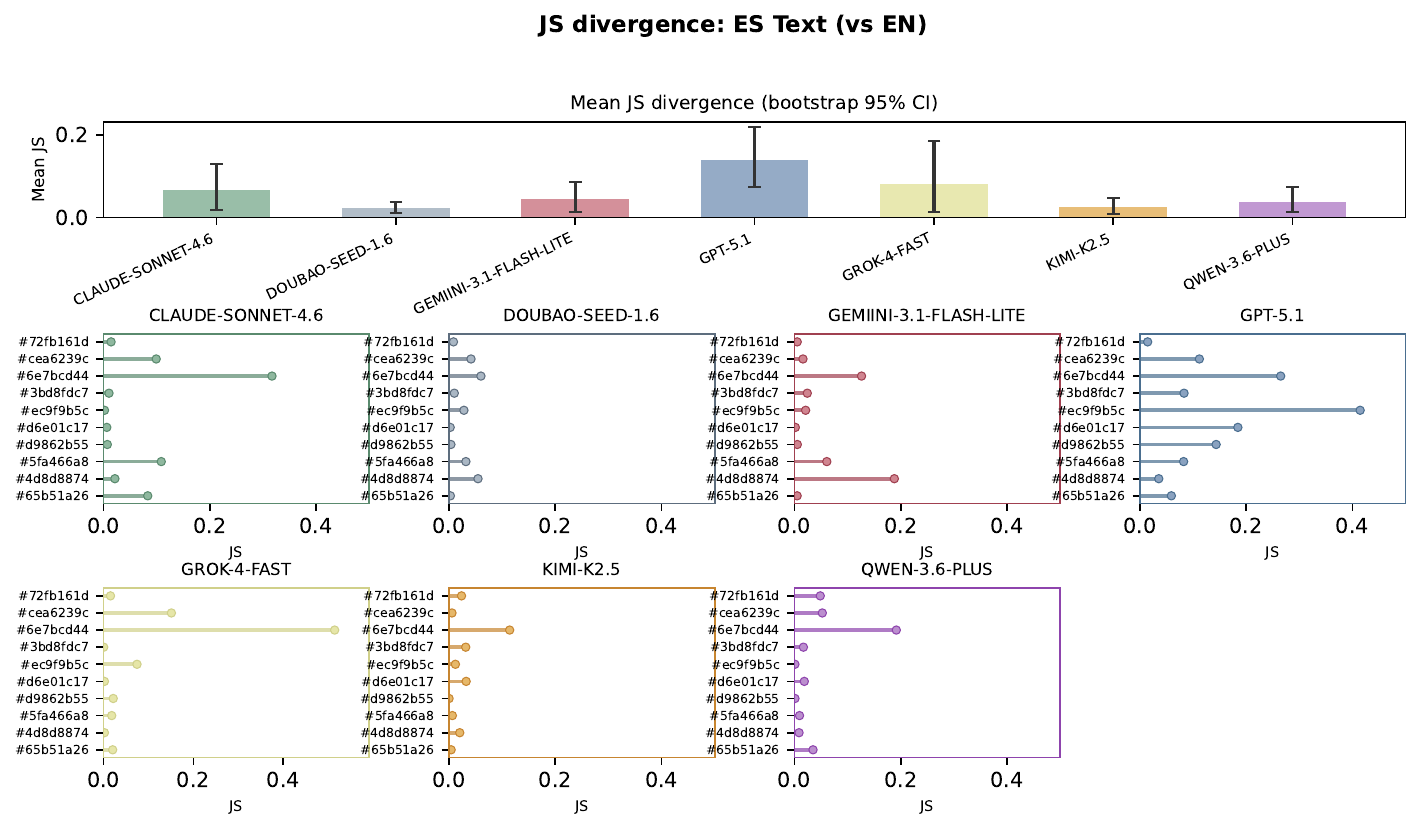}
    \caption{JS divergence: ES Text vs. EN}
    \label{fig:js_es_text}
\end{figure*}

\begin{figure*}[!t]
    \centering
    \includegraphics[width=\textwidth]{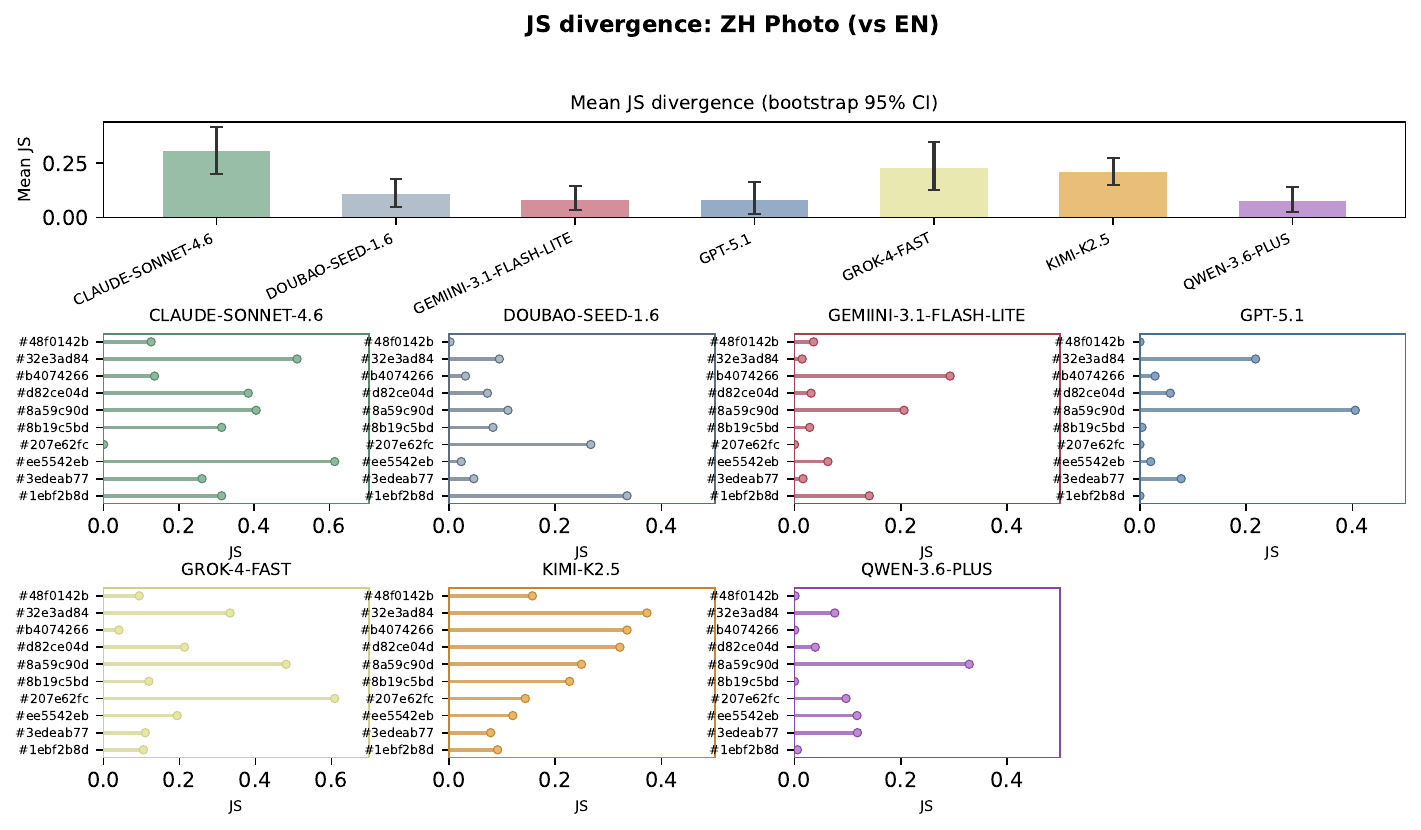}
    \caption{JS divergence: ZH Photo vs. EN}
    \label{fig:js_zh_photo}
\end{figure*}

\begin{figure*}[!t]
    \centering
    \includegraphics[width=\textwidth]{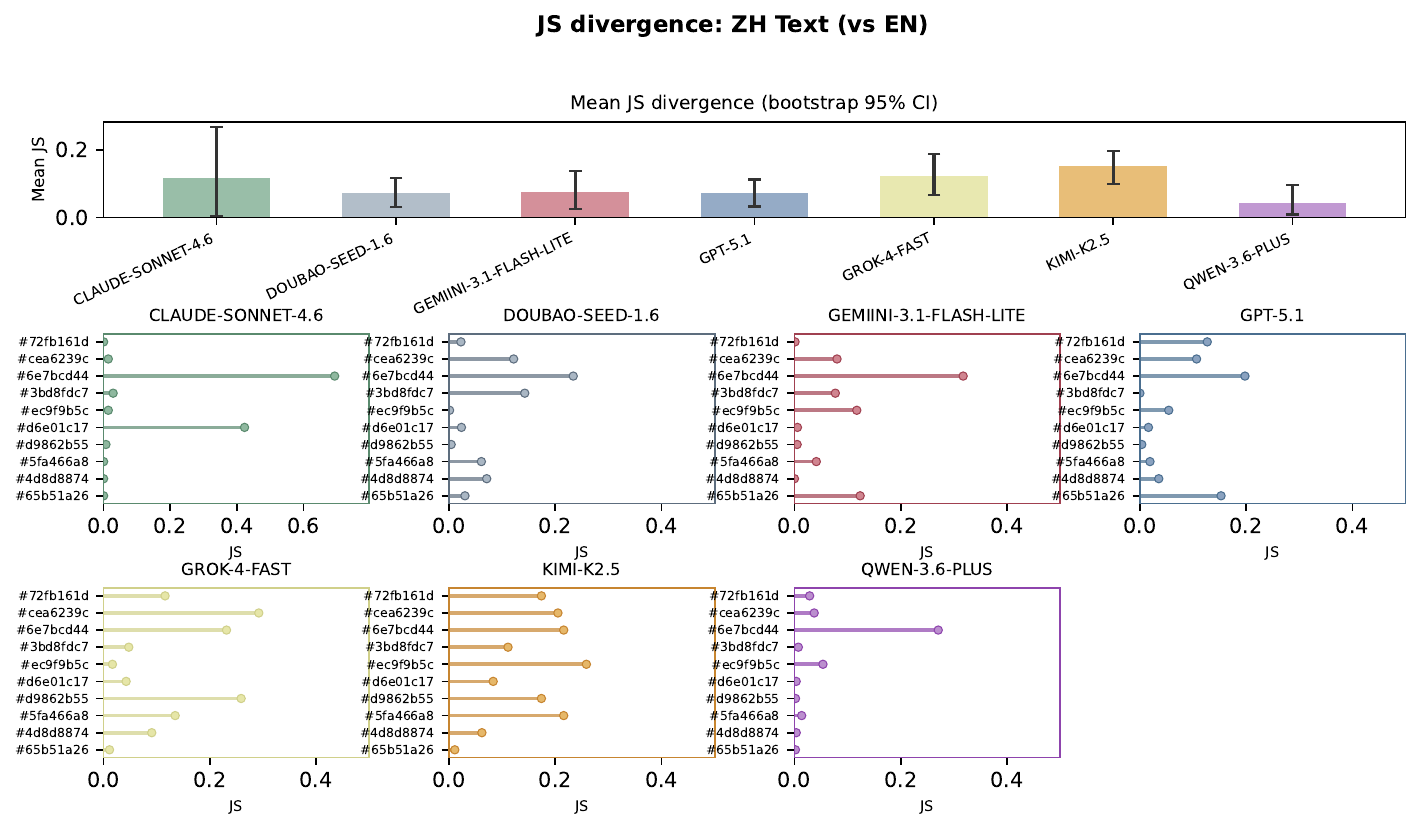}
    \caption{JS divergence: ZH Text vs. EN}
    \label{fig:js_zh_text}
\end{figure*}

\begin{figure*}[!t]
    \centering
    \includegraphics[width=\linewidth]{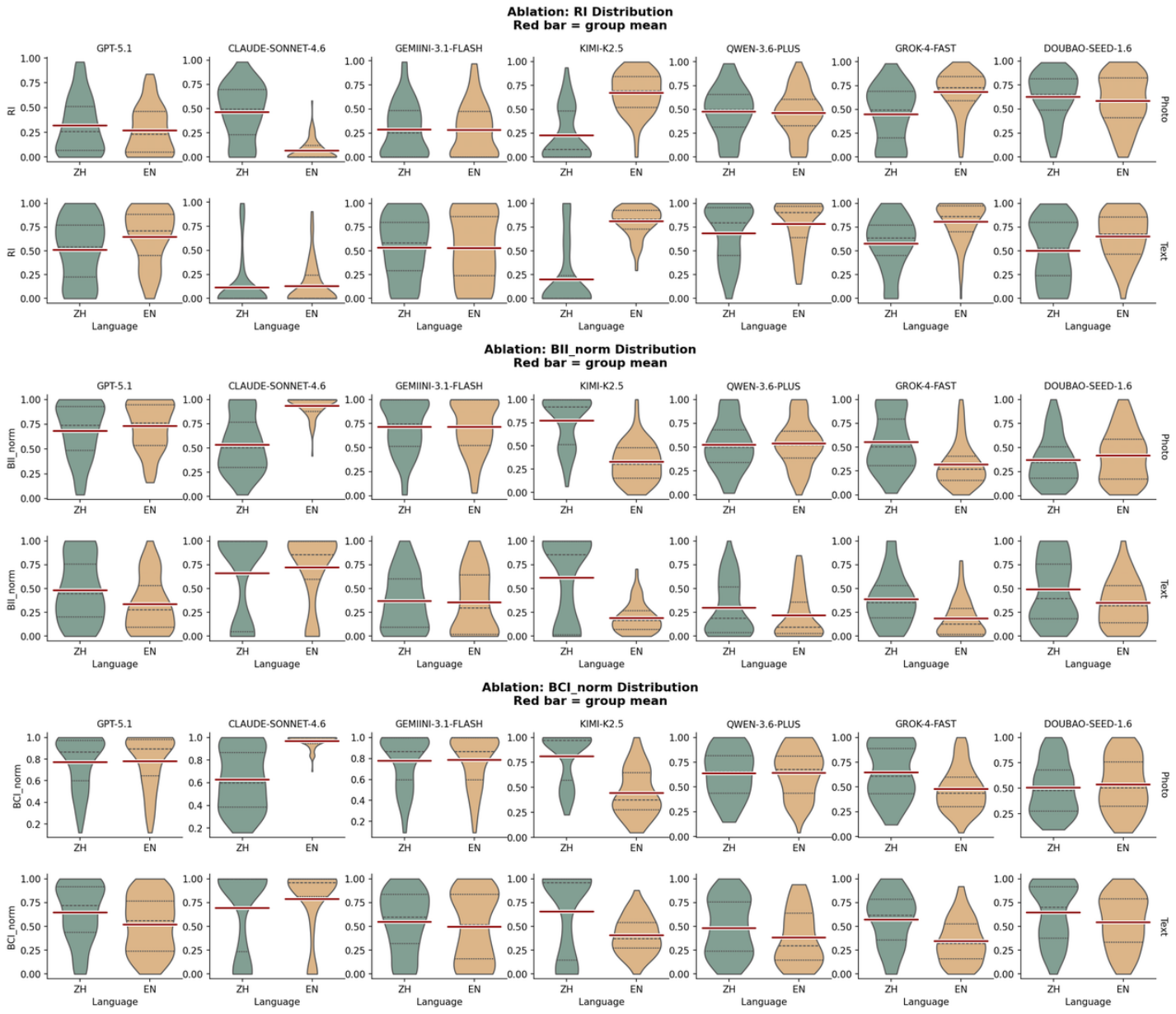}
    \caption{Data distribution between English and Chinese}
    \label{fig:violin}
\end{figure*}

In graphic and object preference tasks, the probability distributions of the seven models exhibited strong language- and corpus-dependent biases across all four languages (Chinese, English, French, and Spanish).
For example, in an equivalent graphic selection task, Claude-4.6 showed severe stochastic collapse under English instructions, assigning 100\% of selections to the ``Triangle'' shape, yet under French instructions its distribution shifted to a completely different option (JS divergence = 0.693). Under Chinese instructions, this polarization weakened, with the distribution shifting toward a mixture of star and triangle preferences. In contrast, Kimi-k2.5 selected ``Green Star'' 50 out of 50 times under Chinese prompts but only 38 times under English, while Qwen 3.6 Plus selected ``Green Star'' approximately 40 times under both languages, indicating that some models exhibit language-conditioned bias shifts while others maintain consistent preferences across languages. Similarly, GPT 5.1 exhibited a substantially different choice distribution under Spanish instructions compared to English for certain visual tasks (JS divergence = 0.536), indicating that language switching strongly modulates its decision manifold even when the model maintains moderate randomness in English. These results suggest that language-conditioned token probabilities directly influence cross-modal feature fusion, producing strongly language-dependent biases that are not limited to any single language pair. Figure~\ref{fig:violin} depicts the aggregate RI  score distributions by model, modality, and language (Chinese and English only). Language-pair-specific JS divergences are broken down by modality in Figures~\ref{fig:js_fr_photo} and \ref{fig:js_fr_text} (French), \ref{fig:js_es_photo} and \ref{fig:js_es_text} (Spanish), and \ref{fig:js_zh_photo} and \ref{fig:js_zh_text} (Chinese).

Spatial geometric preferences and coordinate mappings were also strongly modulated by instructional language, often producing opposite decision patterns within the same model. In the grid-point selection task, English prompts biased selections toward the bottom-left coordinate, whereas Chinese prompts concentrated the probability mass on the top-left and top-right nodes. French and Spanish prompts induced yet other spatial biases, as captured by elevated JS divergence values in the corresponding language-pair plots.

For text-based random choice tasks, similar cross-lingual distributional shifts were observed across all language pairs, with JS divergence values frequently exceeding 0.15, confirming that language-conditioned implicit biases extend beyond visual modalities to purely linguistic selection tasks. Several models exhibit substantial preference inversions solely due to language change.

Overall, the fine-grained JS divergence analysis across all language pairs (Chinese, English, French, Spanish) demonstrates that implicit stochastic collapse is not merely a bilingual phenomenon but a pervasive, language-conditioned property of contemporary MLLMs.

\subsection{Robustness to Label Symbolism}
\label{app:ablation_symbol}

To systematically investigate the underlying mechanics of the implicit subconscious when standard response tokens are neutralized, this section provides an extended quantitative analysis of the symbol-substitution ablation experiments. By isolating the models from conventional alphanumeric identifiers, we observe the raw interaction between the vocabulary embedding space and logic-neutral contexts. The empirical distributions across three distinct non-standard token regimes demonstrate that stochastic collapse is universally preserved. This confirms that the observed deviations are not merely surface-level artifacts of canonical alphabetical ordering, but rather deep-seated structural vulnerabilities within the high-dimensional parameter manifold.

\begin{figure*}[!t]
    \centering
    \includegraphics[width=\textwidth]{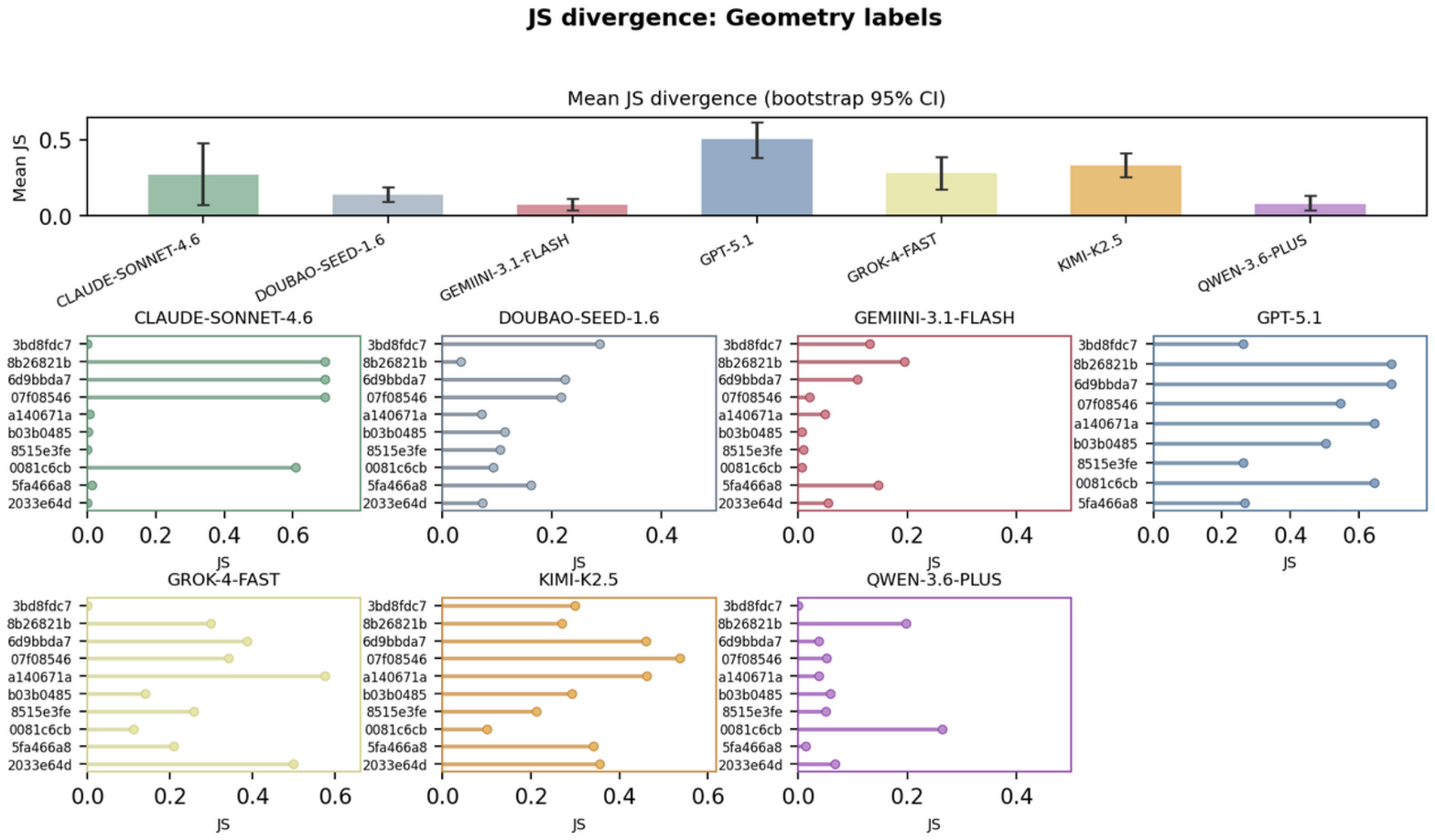}
    \caption{JS divergence: Geometry labels}
    \label{fig:js_geometry}
\end{figure*}

In the \textbf{geometric shape} substitution experiment, the canonical option labels were replaced with structurally symmetric visual primitives ($\blacktriangle$,$\blacksquare$,$\bullet$,$\blacklozenge$). Instead of eliminating semantic preferences to restore a uniform distribution, this substitution induced severe localized distortions and rigid deadlocks on specific geometric tokens, as reflected by the JS divergence distributions in Figure~\ref{fig:js_geometry}. For instance, in logic-neutral card-drawing tasks, Claude-4.6 exhibited an absolute mode collapse by allocating one hundred percent of its empirical probability mass exclusively to the "$\blacksquare$" symbol. Under identical conditions, Gemini 3.1 Flash-Lite demonstrated an alternative but similarly rigid attractor, concentrating seventy-four percent of its responses onto the "$\bullet$" token. Furthermore, GPT 5.1 showed a profound stickiness toward the "$\blacklozenge$" symbol with 35 selections out of 50, while Grok 4 Fast heavily favored the "$\bullet$". These behaviors confirm that geometric primitives act as potent perceptual anchors that forcefully reshape the decision manifold.

\begin{figure*}[!t]
    \centering
    \includegraphics[width=\linewidth]{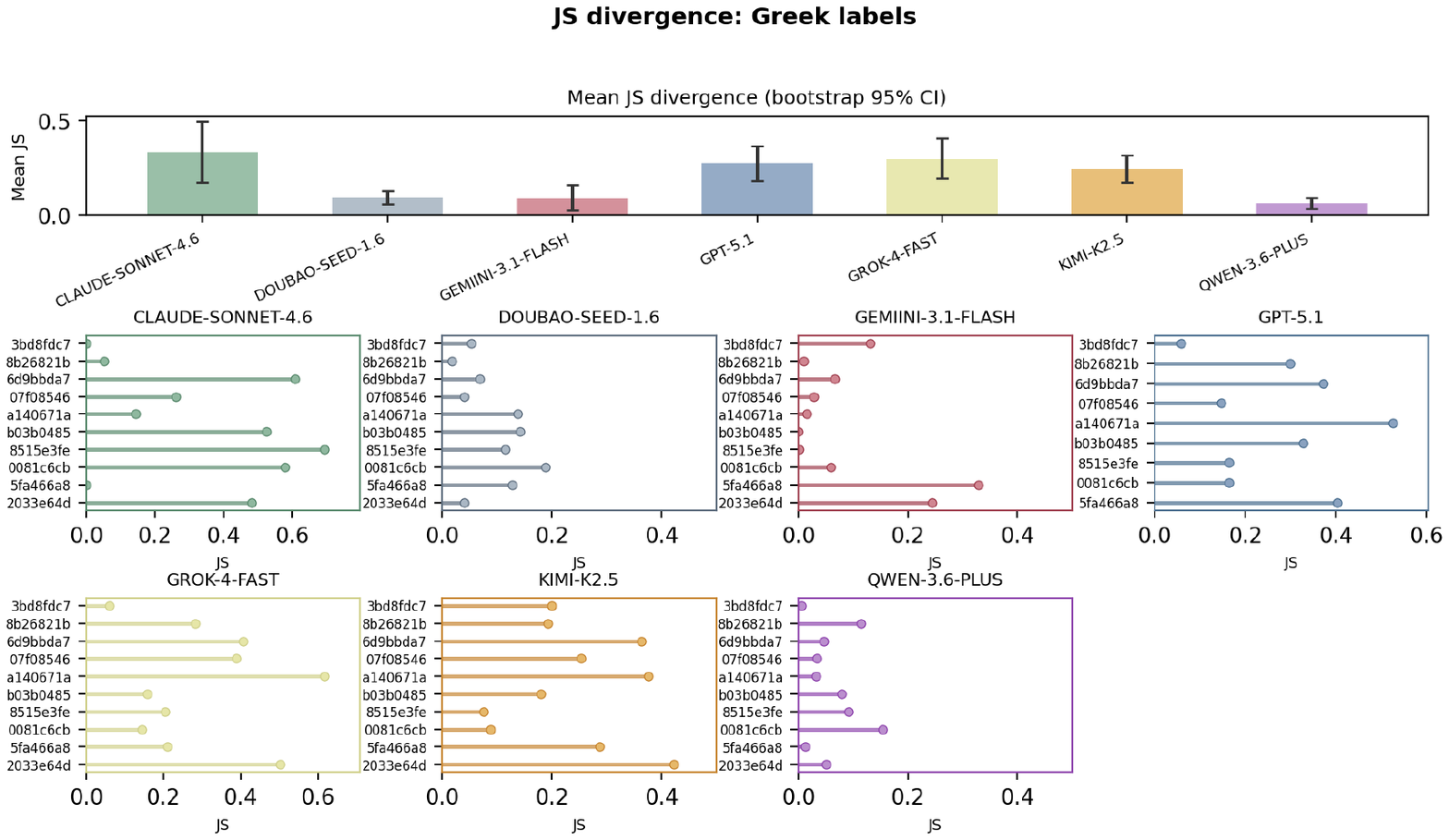}
    \caption{JS divergence: Greek labels}
    \label{fig:js_greek}
\end{figure*}

The introduction of \textbf{Greek alphabet} primitives targeted the models' localized text-frequency priors, which are characterized by highly non-uniform distributions across scientific and linguistic pre-training corpora. This regime revealed a striking cross-model resonance, particularly with the $\gamma$ token, as quantified by Figure~\ref{fig:js_greek}. In repetitive trials within identical contexts, both Claude-4.6 and Grok 4 Fast demonstrated highly convergent parallel deadlocks, allocating 86\% and 72\% of their respective probability masses to the $\gamma$ symbol. This convergent behavior strongly suggests that the high frequency of $\gamma$ as a referential token in human scientific literature has crystallized into a deep-seated cognitive template within the foundational layers of these models. Conversely, GPT 5.1 and Qwen 3.6 Plus exhibited divergent vocabulary biases, with GPT 5.1 predominantly favoring the $\beta$ token and Qwen 3.6 Plus demonstrating a significant 40\% selection rate for the $\alpha$ token. Such divergence underscores that implicit biases are finely modulated by the unique token-frequency topographies of each model's native training mixture.

\begin{figure*}[!t]
    \centering
    \includegraphics[width=\linewidth]{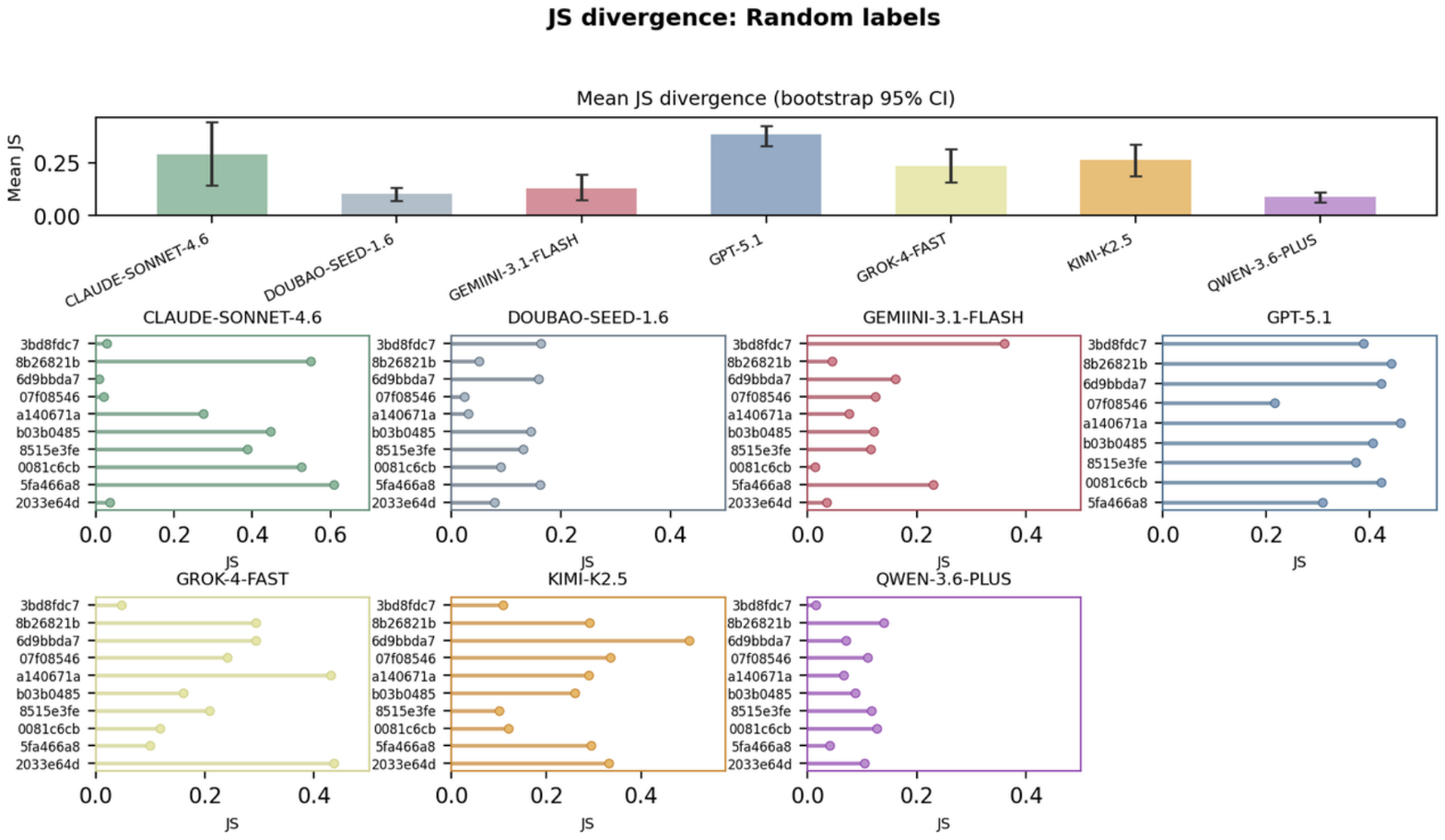}
    \caption{JS divergence: Random labels}
    \label{fig:js_random}
\end{figure*}

To rigorously eliminate all potential human semantic associations and symbol familiarity, we executed an extreme baseline experiment utilizing randomized, \textbf{unique alphanumeric hashes} as option labels. Remarkably, even when the choice candidates were rendered entirely uninterpretable and devoid of natural language logic, the models failed to maintain the maximum entropy baseline of a uniform distribution (see Figure~\ref{fig:js_random} for the JS divergence analysis). In a structurally unconstrained clock-face imagination task, Claude-4.6 concentrated 84\% of its choices on a hash string "tag\_9lifub". Similarly, GPT 5.1 experienced severe stochastic collapse by deadlocking heavily on specific random strings like "tag\_gdgytj", while Grok 4 Fast exhausted 56\% of its responses on the isolated hash "tag\_edybbc", which was significantly higher than the original (A/B/C option) results.

These fine-grained ablation data provide a unified information-theoretic explanation for stochastic collapse, framing it as a phenomenon of fundamental embedding manifold regression under logic-deficient constraints. When the explicit reasoning layer encounters a logic vacuum, it fails to generate definitive gradients or directional attention weights to differentiate the available options. Absent top-down logical guidance, the decoder's forward pass inevitably regresses to rely on bottom-up parameter inertia. Consequently, the final prediction vectors retreat completely to the baseline token-frequency distributions and conditional probability weights ingrained within the pre-trained embedding space. Regardless of whether the token represents a standard letter, a geometric shape, or a randomized hash, the projection layer defaults to the path of least resistance dictated by the highest intrinsic parameter prior. This definitively proves that achieving true cognitive neutrality requires far more than superficial prompt syntax modifications; it demands addressing the data-driven heuristic preferences permanently imprinted into the physical topology of the network's parameters.

\section{Details of Human Expert Review}
\label{appendix:human_review}

\paragraph{Annotator Demographics.} 
The review panel consisted of four independent researchers: three doctoral candidates specializing in Natural Language Processing and one postdoctoral researcher with expertise in Cognitive Science. All reviewers were fully bilingual in English and Chinese. In addition, native French- and Spanish-speaking annotators assisted with multilingual data verification and translation. All reviewers and annotators were compensated at a rate of \$25/hour, which was well above the local minimum wage.

\paragraph{Annotation Instructions.} 
The reviewers were provided with the following unified written guidelines during the quality control phase:
\begin{verbatim}
1. Logic-Neutrality Check
  Verify that all candidate options within 
an instance are strictly logically equiva-
lent and  hold  absolutely no utilitarian
advantage over one another.

2. Safety and Harmlessness Filtering
  Immediately flag and discard any image or
text prompt that inadvertently contains or 
induces sociopolitical stereotypes, offen-
sive content, or personally identifiable 
information (PII).
\end{verbatim}

\section{Additional Statements}

\subsection{Potential Impacts and Ethical Considerations}
While RandomBench provides a rigorous framework for evaluating implicit cognitive biases and stochastic collapse in MLLMs, we must acknowledge potential ethical risks. Theoretically, malicious actors could exploit the systematic biases uncovered in our study (e.g., positional priors or visual anchors) to subtly manipulate model decisions in real-world deployments via adversarial prompt engineering. To mitigate potential harms during dataset construction, all candidate items underwent rigorous logic-neutral filtering to strictly exclude any sociopolitical, demographic, identity-related, or offensive expressions. Furthermore, our findings are intended to facilitate the development of debiasing alignment techniques rather than to exploit model vulnerabilities.

\subsection{License and Ethics Statement}
We confirm that all samples in RandomBench are synthetic and logic-neutral, and the dataset does not contain personal information, human subjects, or sensitive content. We also confirm that all experiments follow the usage policies of the evaluated models and APIs. The dataset and evaluation code will be publicly released for research purposes only.

\subsection{LLM Usage}

We only use ChatGPT to assist with writing guidance, including improving the grammar, clarity, and readability of the text. We do not use AI in coding, research innovation or other tasks. The authors take full responsibility for all content presented in this paper.

\end{document}